%% file: main.tex
\documentclass[twoside,11pt]{article}

\usepackage{blindtext}

\usepackage{lastpage}

\usepackage{amsmath,amsfonts,bm}

\usepackage{rotating}

\usepackage[inline]{enumitem}
\usepackage{xcolor}

\usepackage{algorithm}
\usepackage{algpseudocode}

\usepackage{booktabs}
\usepackage{multirow}
\usepackage{subcaption}

\usepackage{jmlr2e}

\input{main-paper/macros}

\usepackage{lastpage}
\jmlrheading{27}{2026}{1-\pageref{LastPage}}{5/25; Revised
3/26}{4/26}{25-0994}{Samuel Cognolato, Alessandro Sperduti, and Luciano Serafini}

\ShortHeadings{FLAGG: Flexible Autoregressive Graph Generation}{Cognolato, Sperduti, and Serafini}
\firstpageno{1}

\begin{document}

\title{FLAGG: Flexible Autoregressive Graph Generation}

\input{main-paper/authors}

\editor{Qiang Liu}

\maketitle

\input{main-paper/abstract}

\input{main-paper/introduction}

\input{main-paper/relwork}

\input{main-paper/method}

\input{main-paper/implement}

\input{main-paper/experiments}

\input{main-paper/conclusion}

\input{main-paper/acknowledgment}

\bibliography{bibtex/references}

\appendix
\input{appendix/proofs}

\end{document}

%% file: main-paper/macros.tex
\DeclareMathAlphabet{\mathsfit}{\encodingdefault}{\sfdefault}{m}{sl}
\SetMathAlphabet{\mathsfit}{bold}{\encodingdefault}{\sfdefault}{bx}{n}
\newcommand{\tens}[1]{\bm{\mathsfit{#1}}}

\def\verts{{\mathcal{V}}} % vertices
\def\edges{{\mathcal{E}}} % edges

\def\split{{\operatorname{split}}} % split
\def\merge{{\operatorname{merge}}} % merge

\def\gG{{\mathcal{G}}}    % graph
\def\gW{{\mathcal{W}}}    % graph with external edges
\def\adj{{\mathbf{A}}}  % adjacency matrix
\def\fX{{\bm{X}}}       % X node features
\def\fE{{\tens{E}}}     % E edge features
\def\R{{\mathbb{R}}}    % real numbers set
\def\E{{\mathbb{E}}}    % expectation
\def\fy{{\bm{y}}}       % y global features
\newcommand{\KL}{D_{\mathrm{KL}}}   % KL divergence

%% file: main-paper/authors.tex
\author{\name Samuel Cognolato \email samuel.cognolato@phd.unipd.it \\
       \addr Department of Mathematics, University of Padova, Padova, VE, 35121 Italy;\\
       \addr Fondazione Bruno Kessler (FBK), Trento, TN, 38123 Italy
       \AND
       \name Alessandro Sperduti \email alessandro.sperduti@unipd.it\\
       \addr Department of Mathematics, University of Padova, Padova, VE, 35121 Italy;\\
       \addr Fondazione Bruno Kessler (FBK), Trento, TN, 38123 Italy;\\
       \addr Department of Information Engineering and Computer Science, University of Trento, Trento, TN, 38123 Italy
       \AND
       \name Luciano Serafini \email serafini@fbk.eu\\
       \addr Fondazione Bruno Kessler (FBK), Trento, TN, 38123 Italy}

%% file: main-paper/abstract.tex
\begin{abstract}%
    The Deep Graph Generation's panorama spans two extremes: one-shot and sequential models. The former generates nodes and edges jointly, while the latter samples them autoregressively. Each method performs better in different graph domains depending on size and topology, but neither is applicable to all graph categories. For instance, one-shot methods struggle with generating large graphs, while sequential methods underperform on smaller graphs. A possible way to overcome these limitations is to flexibly combine the two methods in a unique system. In this work, we propose the FLAGG (Flexible Autoregressive Graph Generation) framework, which sequentially generates portions of graphs with one-shot models. FLAGG can apply any one-shot model to make it autoregressive, allowing flexibility in choosing the sequential policy. This policy is specified through a stochastic node removal process, which an Insertion Model learns to reverse. We evaluate FLAGG with the DiGress one-shot model on several data sets of different graph sizes and domains. We show that the approach outperforms both one-shot and autoregressive baselines in terms of sampling quality.
\end{abstract}


%% file: main-paper/introduction.tex
\section{Introduction}\label{sec:intro}

Graphs are ubiquitous in many fields of science and technology. Generating new graphs from a reference distribution is a task of paramount importance, for instance, in drug design~\citep{vignac2023digress,huang2023cdgs}. The task is usually framed as having an unknown distribution over graphs $p_\text{data}(G)$, and the goal is to learn a model $p_\theta(G)$ that approximates it. The model should allow sampling and, when possible, evaluating the likelihood of a given graph. The main challenges encountered for this task include handling the discreteness of graphs, the combinatorial explosion of the sample space, and the need for inductive biases in the model. For instance, a probability distribution over graphs should be permutation invariant, meaning that equal probability is assigned to any permutation of the nodes.

The field has seen rapid growth in recent years. One motivation is the introduction of new powerful generative frameworks, namely Denoising Diffusion Probabilistic Models~\citep{ho2020diffusion}, Score-based Models~\citep{song2021scorebased}, and Flow Matching Models~\citep{lipman2023flow}, which were translated from their native task of image generation to graphs. This wave brought many successful instances~\citep{vignac2023digress,jo2022gdss,huang2023cdgs,chen2023efficient}. This category, also comprising Variational Autoencoders (VAE)~\citep{simonovsky2018graphvae}, Normalizing Flows (NF)~\citep{zang2020moflow} and Energy-Based Models (EBM)~\citep{liu2021graphebm}, share a common operative pattern, that is, generating graphs in \textit{one-shot}. This term refers to filling the labels and connections of all nodes jointly in one step. On the other hand, a parallel line of works proposes to build graphs by iteratively adding new nodes and edges in an \textit{autoregressive} manner~\citep{liao2019gran,luo2021graphdf,kong2023grapharm}. This approach is more flexible, allowing for interventions during the generative process~\citep{kong2023grapharm}, and larger graph generation thanks to the better memory management~\citep{davies2023size}. However, poor parallelizability on hardware and worse sampling performance overshadow these advantages.

Lately, a new trend combining the two distinct families has emerged. The idea is to nest powerful one-shot models inside sequential pipelines, hoping to get the best of both worlds~\citep{liao2019gran,cognolato2024ifh,davies2023size}. The appeal stands in making models compositional, where the problem of graph generation is broken down into smaller, more manageable chunks. Removing the limitations posed by the particular factorization of probability may lead to a more flexible design.

Following this trend, we present a recipe for building autoregressive models with nested one-shot architectures named Flexible Autoregressive Graph Generation (FLAGG). The two main components are an Insertion Model, which samples how many new nodes to add at each step, and a Filler Model, which fills in the content of the nodes and the connectivity with the intermediate graph being generated. A one-shot model can actually play the Filler Model's role. The two components are trained to reverse a graph noise process, gradually corrupting data graphs until they follow a known probability distribution. The approach we follow for this work is to remove nodes as a noise process, with the empty graph as the absorbing final state. The flexibility stands in the infinite ways to design this node removal process, which can be customized, for example, in the order and the number of node removals at each step.

A key observation from the flexibility is that fully one-shot and sequential models (generating one node at a time) are the two extremes of a continuum of hybrid solutions. This axis of sequentiality allows the model to be adjusted to the required time and memory constraints and can be tuned as a hyperparameter. This was also shown in our preliminary work~\citep{cognolato2024ifh}, which we extend. We still employ DiGress~\citep{vignac2023digress} but further explore the customization in multiple directions.
First, we refine the IFH model proposed in \citet{cognolato2024ifh} with architectural modifications, additional features, and training techniques. Second, we design a sparse generation for tackling the problem of generating very large graphs like Cora~\citep{prithviraj2008cora}, a citation network of 2810 nodes. For this particular case, we adopt a node selection technique keen to EDGE~\citep{chen2023efficient}.
Finally, we show that the model can be further factorized into smaller modules, each responsible for sampling properties of interest. In this work, we showcase the use of the degree distribution and distance of nodes as properties to be sampled, improving the overall performance in these aspects.

The paper's contributions are summarized as follows:
\begin{enumerate}
    \item FLAGG extends IFH~\citep{cognolato2024ifh} in: \begin{enumerate*}[label=(\roman*)]
        \item allowing all nodes' embeddings to change during the filler step;
        \item taking an additional set of auxiliary input features, e.g., spectral and topological features;
        \item using techniques like weights Exponential Moving Average (EMA) (like in~\citet{ho2020diffusion}) to stabilize the training process.
    \end{enumerate*}
    \item We build a sparser filler step for generating large graphs, e.g., the sizable citation network Cora.
    \item We show that FLAGG can be further factorized into submodules, each responsible for sampling features of interest. This explicit generation is the first step towards a property-aware graph generation.
    \item FLAGG improves over the original at intermediate sequentiality levels on benchmark data sets like QM9, Zinc, Community-Small, Ego-Small, Enzymes, and Ego.
\end{enumerate}

%% file: main-paper/relwork.tex
\section{Related Work}

In this section we give a brief overview of the two main families of graph generation models: one-shot and autoregressive. We also discuss the recent trend of generating large graphs.

\subsection{One-Shot Graph Generation}\label{sec:rel_work/one_shot}
One way to generate a graph is to sample its adjacency matrix. One-shot models express $p_\theta(G)$ as the joint probability distribution over adjacency matrix and features, letting all nodes and edges be sampled at once. This procedure assumes that the number of nodes is known in advance. Empirical sampling from the frequencies of graph sizes of a data set is usually employed, but other strategies are possible.
Generating the adjacency matrix and features has been tackled through latent variable models like Graph Variational Autoencoders (VAE)~\citep{kipf2016variational,simonovsky2018graphvae}, Graph Generative Adversarial Networks (GAN)~\citep{de2018molgan,martinkus2022spectre}, Normalizing Flows~\citep{zang2020moflow} and Diffusion Models for graphs~\citep{vignac2023digress,chen2023efficient}. The common factor is to define a generative process mapping points from a simple, known distribution to the data space. For example, Diffusion Models gradually denoise a corrupted graph structure starting from a random one.
Score-based models for graphs~\citep{jo2022gdss,huang2023cdgs} use Stochastic Differential Equations (SDE) to model a continuous flow from the adjacency matrix distribution to the known prior distribution and vice versa.

\subsection{Autoregressive Graph Generation}\label{sec:rel_work/autoregressive}
Another way to generate graphs is by sequentially adding new nodes and edges to a growing graph. One then must define a node ordering as a permutation $\pi$. This is the approach taken by autoregressive models.
Some works~\citep{you2018graphrnn,liao2019gran} focus on the architecture of the model in itself, using Recurrent Neural Networks (RNN) or particular Graph Neural Networks (GNN) to compute edge probabilities. Others adopt popular generative frameworks like Normalizing Flows and Diffusion Models in the autoregressive setting~\citep{shi2020graphaf,luo2021graphdf,kong2023grapharm}.
Node ordering has been a hot topic~\citep{chen2021order} and is one of the main criticisms of autoregressive models. Because we usually cannot find a canonical order of nodes, these models must be trained on many permutations, possibly $n!$, to learn permutation invariance. \citet{chen2021order} was the first to introduce the idea of learning the node ordering, showing empirically that results are usually better compared to prefixed orderings such as BFS (Breadth-First Search) or DFS (Depth-First Search). GraphARM~\citep{kong2023grapharm} expanded the idea of learning the ordering through a node absorbing state diffusion model.

\subsection{Generating Large Graphs}\label{sec:rel_work/large_graphs}
Recently, there has been a growing interest in modeling larger-scale generation. One of the main challenges gatekeeping the task is the memory and time constraints. One-shot models have to store the whole graph in memory, which scales quadratically with the number of nodes, making their application unfeasible. For this reason, autoregressive models~\citep{liao2019gran} have been at the forefront of large graph generation, as they can break down the generation process into smaller steps.
Another recent trend that is picking up in pace is that of hierarchical graph generation. The idea is to generate a graph through a series of expansion operations, which gradually increase the resolution of nodes. Recently, the work of~\citet{davies2023size} proposed a hierarchical graph generative model based on DiGress~\citep{vignac2023digress}, which can reconstruct large graphs like Cora~\citep{prithviraj2008cora}.

%% file: main-paper/method.tex
\section{Method}

In this section, we give a detailed explanation of the methodology, integrated with the novel contributions. FLAGG is an extension of our preliminary work~\citep{cognolato2024ifh}.

\subsection{Notation and Definitions}\label{sec:method/defs}
Let $\gG=(\verts,\edges)$ be a graph with nodes \mbox{$\verts=\{v_1,\dots,v_n\}$} and edges $\edges\subseteq\verts\times\verts$. The graph's connectivity can also be represented with an adjacency matrix $\adj\in\{0,1\}^{n\times n}$, where $\adj_{i,j}=1$ if $(v_i,v_j)\in\edges$. If the graph is undirected, $\adj$ is symmetric, i.e., $\adj=\adj^\top$. Labels can be assigned to nodes and edges. In this case we will refer to node labels as \mbox{$\fX\in\{0,1\}^{n\times d_x}$}, and edge labels as $\fE\in\{0,1\}^{n\times n\times d_e}$, represented as tensors of one-hot encoded vectors. Global features of the graph are denoted as $\fy\in\R^{d_y}$. We start our definitions with the concepts of node removal and induced subgraphs.

\begin{definition}[Node removal]\label{def:remv_op}
Removing a node $v_i$ from $\gG$, denoted as $\gG-v_i$, returns a new graph where $v_i$ is removed from the set of vertices $\verts$, together with all edges incident to $v_i$ removed from $\edges$. Removing multiple nodes at once is denoted as $\gG-\verts'$, where $\verts'\subseteq\verts$.
\end{definition}

\begin{definition}[Induced subgraph]\label{def:induced}
The subgraph $\gG'$ induced in $\gG$ by $\verts'\subseteq \verts$ is the subgraph with nodes $\verts'$ and edges $\edges'=\edges\cap \verts'\times\verts'$. It is denoted as $\gG[\verts']$.
\end{definition}

\begin{figure*}[t]
    \centering
    \includegraphics[width=0.75\textwidth]{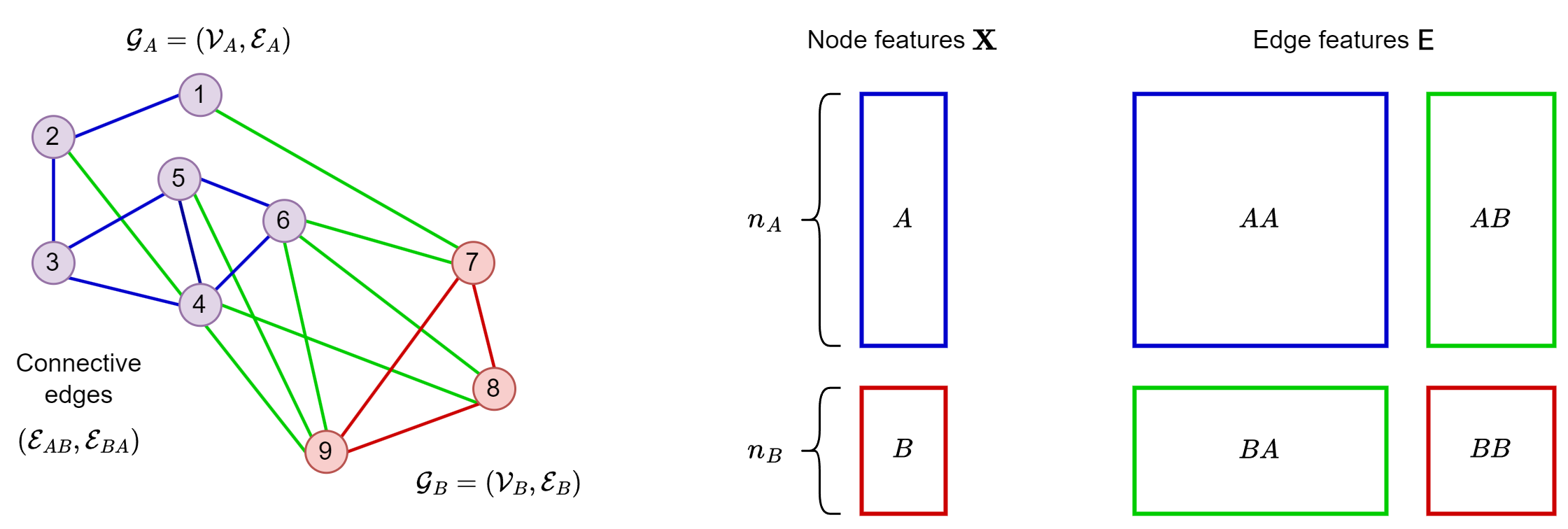}
    \caption{Split operation. The induced subgraphs $\gG^A$ and $\gG^B$ are blue and red. In green are the intermediate edges $\edges^{AB},\edges^{BA}$. On the right is the split adjacency matrix, which has the same coloring.}\label{fig:split_op}
\end{figure*}

We can observe that the removal of nodes is equivalent to the induction of subgraphs, as $\gG-\verts'=\gG[\verts\setminus\verts']$. From the notion of node removal, it follows that if there is a labeling $\fX,\fE$ of nodes and edges, the entries for $v_i$ in $\fX$ and all its edges in $\fE$ are removed. On the opposite side, adding new nodes and edges with labels will concatenate their values to existing nodes and edge labels. To keep the notation simple, we acknowledge this fact now and do not mention it again.
Now, we introduce the graph splitting and merging operations, shown in Fig.~\ref{fig:split_op}, enabling us to define the forward and reversed node removal sequences for FLAGG. One can observe that the two operations are one the inverse of the other.

\begin{definition}[Split operation]\label{def:split_op}
The operation $\split(\gG,\verts^A)$ of $\gG=(\verts,\edges)$ through $\verts^A\subseteq\verts$ and $\verts^B=\verts \setminus \verts^A$ returns a four-tuple $(\gG^A,\gG^B,\edges^{AB},\edges^{BA})$ with:
\begin{equation*}
    \begin{gathered}
    \gG^A=\gG[\verts^A],\quad \gG^B=\gG[\verts^B], \\
    \edges^{AB}=\edges\cap(\verts^A\times\verts^B),\quad \edges^{BA}=\edges\cap(\verts^B\times\verts^A).
    \end{gathered}
\end{equation*}
\end{definition}

\begin{definition}[Merge operation]\label{def:merge_op}
The operation $\merge(\gG^A,\gG^B,\edges^{AB},\edges^{BA})$ of two disjoint graphs $\gG^A,\gG^B$ and their intermediate edges $\edges^{AB},\edges^{BA}$ returns a new graph $\gG$ with $\verts=\verts^A\cup \verts^B$ and $\edges=\edges^A\cup \edges^B\cup \edges^{AB}\cup \edges^{BA}$.
\end{definition}

During the explanation of the insertion process of FLAGG, we will also denote the merge operation as $\merge(\gG^A,\gW^{AB})$, with $\gW^{AB} = (\gG^B,\edges^{AB},\edges^{BA})$, making the notation lighter. $\gW^{AB}$ can be considered a new part of the graph being added to $\gG^A$. Now, we define the principal object used in FLAGG, namely the forward and reversed removal sequences.

\begin{definition}[Forward and reversed removal sequence]\label{def:remv_seq}
A sequence of graphs $\gG_{0:T}={(\gG_t)}_{t=0}^T$ is a forward removal sequence of $\gG$ if $\gG_0=\gG$, $\gG_T$ is the empty graph $\varnothing$, and $\gG_t=\gG_{t-1}[\verts_t]$ for some $\verts_t$. A removal sequence can be reordered in reverse to have $\gG_{T:0}={(\gG_t)}_{t=T}^0$, which is called a reversed removal sequence.
\end{definition}

We denote $\mathcal{F}(\gG,T)$ and $\mathcal{R}(\gG,T)$ as the sets of all forward and reversed removal sequences of $\gG$ of length $T$.

\subsection{Removing Nodes as a Graph Noise Process}\label{sec:method/remv_intro}

Diffusion Models are a staple in modern generative models and have been successfully applied to graphs. Still, not every aspect of graphs has been captured by Diffusion Models. Most prominently, the size of a graph needs to be sampled before running a one-shot model. The FLAGG framework incorporates this missing piece into the mix, defining a diffusion model over the whole graph structure. In this light, a diffusion process will manipulate the graph structure, potentially varying its size along the trajectory. In this work, we define it as a node removal process, which gradually corrupts the graph structure until it collapses into an empty graph.

A node removal process is defined as a Markov Process which, given a graph $\gG_{t-1}$ at time $t-1$, samples a subgraph $\gG_t$ at time $t$ by removing a set of nodes $\verts_{t}^{B}\subseteq \verts_{t-1}$ from $\gG_{t-1}$, i.e., $\gG_t=\gG_{t-1}-\verts_{t}^{B}$ (see Definition~\ref{def:remv_op}).
The transition probability of the process is given by $q(\gG_t|\gG_{t-1})$.
Repeatedly applying the node removal process on a graph $\gG$ yields a forward removal sequence $\gG_{0:T}=(\gG_t)_{t=0}^T$ (see Definition~\ref{def:remv_seq}). The probability of the forward removal process, given a starting point $\gG_0=\gG$, is:
\begin{equation}\label{eq:remv_proc}
    q(\gG_{1:T}|\gG_0)=\prod_{t=1}^T q(\gG_t|\gG_{t-1}).
\end{equation}
The key insight for this autoregressive framework is that the removal transition can be factorized as:
\begin{equation}\label{eq:broken_remv}
    \begin{aligned}
        q(\gG_t|\gG_{t-1}) &= q(\gG_t,n_t|\gG_{t-1}) \\
        &= q(\gG_t|n_t,\gG_{t-1})q(n_t|\gG_{t-1}),
    \end{aligned}
\end{equation}
where $q(n_t|\gG_{t-1})$ is the probability that $\gG_t$ will have exactly $n_t$ nodes, and $q(\gG_t|n_t,\gG_{t-1})$ is the probability of obtaining $\gG_t$ by choosing $n_t$ nodes among those in $\verts_{t-1}$ to keep. In other words, $q(n_t|\gG_{t-1})$ tells \textit{how many} nodes are kept alive, while $q(\gG_t|n_t,\gG_{t-1})$ determines \textit{which} $n_t$ nodes are selected. This factorization is possible by observing that, among all possible numbers of nodes, the only one having a nonzero probability $q(\gG_t|n_t,\gG_{t-1})$ is for $n_t=|\verts_t|$.

This definition makes no assumptions about the type of graphs or the removal policy. Due to this flexibility, one can design any removal policy, even one tailored to the specific hardware constraints or graph structure. The Absorbing Node State diffusion seen in GraphARM~\citep{kong2023grapharm} shares some similarities with the above definitions, where masking replaces removal. We see it as an instance of the more general framework where the ordering of nodes is learned, and the number of removed nodes is always 1. We touch upon this point again in Section~\ref{sec:inst/spec}.

\subsection{Parameterizing the Reverse Removal Process}\label{sec:method/rev_proc}

As with Diffusion Models, a neural network has to learn to reverse the removal process defined in the previous section.
Reversing a node removal process consists in the introduction of new nodes and edges. This is not trivial, as the specific implementation depends on the nuances of the particular way of removing nodes. In this section, we give the general definition of the reverse process, how to parameterize it, and how to learn its parameters.

The reverse process aims to recover a graph $\gG$ corrupted by a removal process, starting from an empty graph. Insertion is done by adding new portions of the graph to the growing graph across an appropriate amount of time steps. This is formalized by a node insertion process which, given a partial graph $\gG_t$, samples a new graph $\gG_{t-1}$ with $n_{t-1}\geq n_t$ nodes, and $\gG_t$ being its subgraph. This step requires to sample a new subgraph $\gG_{t}^{B}=(\verts_{t}^B,\edges_{t}^B)$ with $r_t=n_{t-1}-n_t=|\verts^B|$ nodes, and edges $\edges_{t}^{AB},\edges_{t}^{BA}$ to connect it to $\gG_t$. Then $\gG_{t-1}$ is obtained through a merge operation between $\gG_{t}^A=\gG_t$ and the new portion $\gW_{t}^{AB}=(\gG_{t}^B,\edges_{t}^{AB},\edges_{t}^{BA})$ (see Definition~\ref{def:merge_op}). From now on, we use $\gW_t$ instead of $\gW_{t}^{AB}$ for brevity. Starting from the empty graph $\varnothing$, we obtain the reversed removal sequence $\gG_{T:0}=(\gG_t)_{t=T}^0$ (see Definition~\ref{def:remv_seq}). The insertion transition $p_{\Theta}(\gG_{t-1}|\gG_t)$ models one step of it, where $\Theta$ are the model's parameters. The reverse process is defined as:
\begin{equation}\label{eq:ins_proc}
    p_{\Theta}(\gG_{T:0})=p_{\Theta}(\gG_T)\prod_{t=1}^T p_{\Theta}(\gG_{t-1}|\gG_t).
\end{equation}
Now, this form includes a distribution for $\gG_T$, but in the case of the empty graph $\varnothing$, the factor $p_{\Theta}(\gG_T)$ can be removed as all mass is placed on $\varnothing$.
As for the removals, also the insertions can be factorized with the same trick:
\begin{equation}\label{eq:broken_ins}
    \begin{aligned}
        p_{\Theta}(\gG_{t-1}|\gG_t) &= p_{\Theta}(\gG_{t-1},r_t|\gG_t)\\
        &= p_{\Theta}(\gW_{t},r_t|\gG_t)\\
        &= p_{\theta}(\gW_{t}|\gG_t,r_t)p_{\phi}(r_t|\gG_t).
    \end{aligned}
\end{equation}
In this context, the Insertion Model $p_{\phi}(r_t|\gG_t)$ samples the number of nodes to add, and the Filler Model $p_{\theta}(\gW_{t}|\gG_t,r_t)$ generates the content of the new nodes and edges, and how to connect them to $\gG_t$. The parameters $\Theta=(\theta,\phi)$ are learned by minimizing the variational upper bound, detailed in~\citet{cognolato2024ifh}. Briefly, the adopted loss to minimize is:
\begin{equation}\label{eq:ifh_loss}
    \begin{aligned}
        L_\text{vub}&=\E_{\gG_0\sim q(\gG_0)}\Bigg[
        \sum\limits_{t=2}^{T}\KL \big(q(r_t|\gG_t,\gG_{0}) \Vert p_{\phi}(r_t|\gG_t)\big) + \\
        &+\sum\limits_{t=2}^{T}\KL \big(q(\gW_{t}|\gG_t,r_t) \Vert p_{\theta}(\gW_{t}|\gG_t,r_t)\big)+ \\
        &-\E_{\gG_1\sim q(\gG_1|\gG_0)}\left[\log p_\phi(r_1|\gG_1) + \log p_\theta(\gW_1|\gG_1)\right]
        \Bigg].
    \end{aligned}
\end{equation}
As with Diffusion Models~\citep{ho2020diffusion} and Score-based Models~\citep{song2021scorebased}, the model learns to recover the true data distribution, and not to reconstruct specific data points given their corrupted counterparts. In our case, the generative process starts from $\varnothing$, and expands its support through a series of node additions. As the process evolves, the graphs in the support should resemble subgraphs from the data set.

Another key observation about FLAGG is that $p_{\theta}(\gW_t|\gG_t,r_t)$ can be modeled as any one-shot model, as long as it can also generate the interconnections between the new nodes and the partial graph $\gG_t$. We detail how this can be done in Section~\ref{sec:inst/adapting}. This takes FLAGG at the forefront of recent trends of nesting one-shot models in a modular framework, like in HIGGs~\citep{davies2023size} and SaGess~\citep{limnios2024sagess}.

\subsection{Controlling Graph Growth through Removal Processes}\label{sec:method/remv_procs}
The framework of FLAGG simplifies the design of the generative process by choosing a formulation of $q(\gG_t|\gG_{t-1})$. If we decide how nodes are removed, we also get how the model should add them back. The key principles to follow are:
\begin{enumerate}
    \item The removal process should be reversible. This means that one can find an expression for the reversed process that can be plugged into the model's loss in Equation~\eqref{eq:ifh_loss}. This lets us train a neural network to approximate the process, thus allowing us to sample from it.
    \item The maximum and variance of the number of removed nodes directly impact the memory and time complexities of the algorithms. Control over these quantities is crucial as the filler model needs to materialize the whole adjacency matrix of the new/removed nodes. For example, having high variance and high maximum block size leads to heavy use of padding when batching many adjacency matrices together;
    \item Controlling the order in which nodes are removed can positively impact the learning process, as the model can leverage the graph's structure. For example, we can enforce having a single connected component during removals, which translates into a model that tends to connect new subgraphs to the main body.
\end{enumerate}
Thanks to the factorization in Equation~\eqref{eq:broken_remv}, we can explore the design of $q(n_t|\gG_{t-1})$ as the linkage of nodes is taken care of by the Filler Model. A quantity of notice is the \textit{level of sequentiality}, which we define as the average number of nodes removed at each step, controlling the rate at which the graph grows.
We describe two removal processes: the first randomly and independently selects nodes to remove; the second removes blocks of predefined sizes, overcoming the limitations posed by the use of the former naive process. We detail the two methods in the following, and discuss node ordering later. We will refer to $r_t$ as the number of nodes removed at step $t$, and to $\Delta n_t=n_0-n_t$ as the number of nodes removed over $t$ steps. All proofs for this section can be found in the supplementary material.

\subsubsection{Naive/Binomial}
A naive way to approach node removals is to choose, for each node, whether to keep it or remove it with some probability. Formally, we can model the events of removing the nodes of $\gG_{t-1}$ as Bernoulli random variables, each with probability $q_t$. Counting the number of removals, we obtain a Binomial random variable $B(r_t;n_{t-1},q_t)$. The forward removal transition is then:
\begin{gather}
    q(r_t|\gG_{t-1})= B(r_t;n_{t-1},q_t) \label{eq:bin_next_num}
\end{gather}
Iterating the process many times, starting from $\gG_0$, is equivalent to repeating these Bernoulli removals. When marginalized over the intermediate steps, $\Delta n_t$ is distributed as:
\begin{gather}
    q(\Delta n_t|\gG_{0})= B(\Delta n_t; n_{0}, 1-\pi_t) \label{eq:bin_tsteps_num} \\
    \text{with }\pi_t = \prod_{k=1}^t(1-q_k) \nonumber
\end{gather}
The target of training, the posterior $q(\gG_{t-1}|\gG_{t},G_{0})$, can be obtained using Bayes' rule and is distributed as:
\begin{gather}
    q(r_t|\gG_{t},\gG_{0})=q(r_t|\Delta n_t) = B(r_t; \Delta n_t, 1-\bar{q}_t) \label{eq:bin_post_num}\\
    \text{with }\bar{q}_t =1-\frac{1-\pi _{t-1}}{1-\pi _{t}}. \nonumber
\end{gather}
Intuitively, this makes sense, as the Binomial models how many nodes $r_t$ to add back among the removed $\Delta n_t$ nodes.
The insertion process can be parameterized as:
\begin{equation}
p_\phi(r_t|G_{t})=\sum_{\Delta n_{t}=0}^{\infty}q(r_t|\Delta n_t)p_\phi(\Delta n_t|\gG_{t}),
\end{equation}
where $p_\phi(\Delta n_t|G_{t})$ predicts the number of missing nodes from $\gG_t$. In IFH~\citep{cognolato2024ifh}, it was defined as a neural network predicting the value $\Delta n_t$. Then, the KL divergence of Equation~\eqref{eq:ifh_loss} is minimum when $p_\phi(\Delta n_t|G_{t})$ correctly predicts the true missing nodes, which corresponds to minimizing the Mean Square Error (MSE).
This strategy has been evaluated in~\citet{cognolato2024ifh}, and presents major drawbacks regarding memory and time complexity. This is motivated by the high maximum and variance of the number of nodes caused by the binomial distribution, leading to a great waste of memory.

\subsubsection{Categorical}\label{sec:method/remv_procs/cat}
One way to have a well-behaved distribution over the number of nodes to insert/remove is to restrict the choice of $r_t$ to a small, finite set of $c$ possibilities (block sizes) $D=~\{d_1,\ldots,d_c\}\subset\mathbb{N}$.
The removal process will then select the number of nodes to remove from $D$ such that their sum results in $n_0$. A removal transition is therefore modeled as a categorical distribution over these block sizes:
\begin{equation}
    q(r_t|\gG_{t-1})=\operatorname{Cat}(r_t;D,q_t),
\end{equation}
where $q_t$ are the probabilities over the elements of $D$.
Since we want to minimize the number of generation steps and the memory consumption, we have to find the minimal set of blocks that sum up to $n_0$.
The optimal solution can be found in solving the change-making problem~\citep{wright1975change}. The number of nodes is interpreted as the amount to return using as few coins as possible from a given set of denominations $D$. This problem can be solved in pseudo-polynomial time using dynamic programming. Any permutation of the coins in a solution gives us the shortest possible removal trajectories $\gG_{0:T}$ with the possibilities in $D$. In particular, the number of steps $T$ will always be the number of coins that make up the amount $n_0$. The process of picking a random coin from the sequence is Markovian, as we only care about the current number of nodes. This is because when we choose a coin $r_t$ and remove it from $n_{t-1}$, the remaining coins are still the solution of $n_t=n_{t-1}-r_t$. The probabilities $q_t$ are then proportional to the histogram $h(n_{t-1})$ of occurrences of each denomination in the sequence given by $n_{t-1}$. The categorical transition is defined as:
\begin{equation}
    q(r_t=d|G_{t-1})=\frac{h(n_{t-1})[d]}{T-(t-1)},
    \label{eq:cat_next_num}
\end{equation}
where $h(n_{t-1})[d]$ is the histogram value for denomination $d$.
Sampling a denomination can be seen as extracting a colored ball from an urn, where $h(n_{t-1})$ are the number of balls for each color, determining the denominations. Iterating this process from $\gG_0$ can be seen as extracting multiple balls from the urn, which starts at $h(n_0)$, and every extraction removes a ball. It can be recognized that the resulting counts of colored balls over $t$ extractions are distributed as a multivariate hypergeometric distribution. The $t$-step marginal is then:
\begin{equation}
    q(\Delta n_{t}|G_0)=\frac{\prod_{d\in D}\binom{h( n_{0})[d]}{h(\Delta n_{t})[d]}}{\binom{T}{t}}, \label{eq:cat_tsteps_num}
\end{equation}
where, again, $\Delta n_t=n_0-n_t$ is the number of missing nodes between $\gG_0$ and $\gG_t$. Reversing the process yields the posterior:
\begin{equation}
    q(r_t=d|G_{t},G_{0})=q(r_t=d|\Delta n_t)=\frac{h(\Delta n_t)[d]}{t}.
    \label{eq:cat_post_num}
\end{equation}
Notice that the posterior is precisely the same as the forward transition, considering the amount $\Delta n_t$ instead of $n_t$.
The task of the insertion model is to predict the distribution of missing nodes $h(\Delta n_t)/t$, and this can be done by minimizing the KL divergence between the posterior and the predicted distribution. We normalize the target into probabilities to avoid learning the scale (i.e., the number of nodes). This process optimizes memory usage by having larger blocks with less padding and reduces time by minimizing the number of steps.

\subsubsection{Node Ordering}\label{sec:method/remv_procs/order}
Other than controlling \textit{how many} nodes are removed, one can also control \textit{which} nodes are removed, as described in Equation~\eqref{eq:broken_remv}. Either this is directly implemented in the removal process, like the naive one, which is uniform over all nodes, or it can be predefined before starting the removal process. Particular node orderings can give useful inductive biases to reduce the training examples needed to learn permutation equivariance~\citep{liao2019gran,chen2021order}. For example, Breadth First Search (BFS) starts from a random root node and orderly visits neighbors of distance 2, then 3, and so on, which we call layers. Then, the removal process will start from the outermost layer and proceed inwards to the root. BFS ensures that a graph $\gG_t$ always has one connected component. In general, given a node ordering $\pi$, the transitions will be of the form:
\begin{equation}
    \begin{aligned}
        q(\gG_t|\gG_{t-1},\pi) &= q(\gG_t|n_t,\gG_{t-1},\pi)q(n_t|\gG_{t-1},\pi) \\
        &= q(n_t|\gG_{t-1},\pi).
    \end{aligned}
\end{equation}
where $q(\gG_t|n_t,\gG_{t-1},\pi)$ deterministically selects the first $n_t$ nodes in the order as $\gG_t=\gG_{t-1}[v_{\pi(1)},\dots,v_{\pi(n_t)}]$. In this case, the design of a removal process can be broken down into choosing a node order and a process on the number of nodes to remove.

\begin{figure*}[t]
\begin{center}
\includegraphics[width=1.\textwidth]{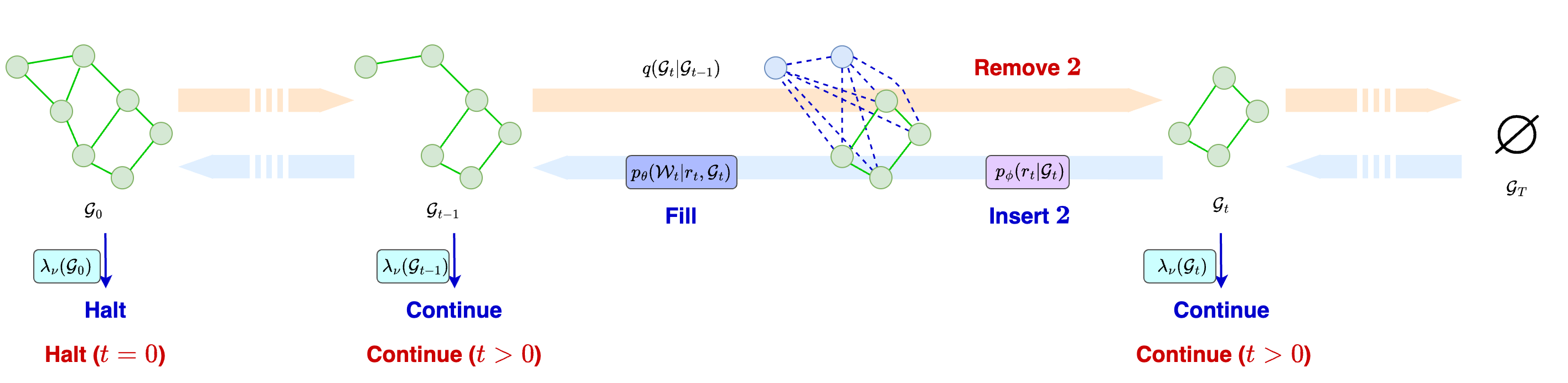}
\end{center}
\caption{How FLAGG works: during training, a graph is corrupted (left to right) by iteratively removing nodes until the empty graph $\varnothing$ is left. At each step, the insertion (violet), filler (blue), and halt (cyan) models have to predict how many nodes were removed, what content they had, and whether the graph is terminal, respectively (right to left).}\label{fig:ifh}
\end{figure*}

\subsection{Halting Process}\label{sec:method/halt}
Graphs are inherently arbitrary in size, and removal processes should account for this property to be supported. A fixed number of steps will result in nonuniform block sizes $r_t$ if the data set presents a large variability in graph sizes. This can be avoided by adapting the number of steps $T$ to the size of a graph, i.e., taking larger $T$ for larger graphs. The model must infer when to stop the generative process to account for this adaptive generation time. This is already integrated into Insertion Models that predict the number of missing nodes $\Delta n_t$. Still, this training objective is hard, as the model tends to learn the average of $\Delta n_t$~\citep{cognolato2024ifh}. For the Insertion Models that do not include this mechanism, e.g., the Categorical insertion, we must include a Halting Model, predicting the probability of a partial graph being a terminal graph. This model is trained in a binary classification setup, where the halting ground truth signal is set to $1$ for the data set graphs and $0$ for all their proper induced subgraphs. This Halting Process has been formalized in \citet{banino2021pondernet}, defining a Markov process $\Lambda_t$ over two states: \textit{continue} ($0$) and \textit{halt} ($1$). The process starts in $0$, and with probability $\lambda_t$ jumps to the absorbing state $1$. Halting can be integrated into the insertion model as sampling zero as the number of added nodes. By defining $\lambda_{\phi_2}(\gG_t)=p_{\phi}(r_t=0|\gG_t)$, we can write the probability of halting exactly in $T$ steps as:
\begin{equation}
    p_{\phi}\left(\Lambda_{0}=1,\Lambda_{T:1}=\mathbf{0}|\gG_{T:0}\right)=\lambda_{\phi_2}(\gG_0)\prod_{t=1}^{T}(1-\lambda_{\phi_2}(\gG_t)).
\end{equation}
In this work, we split the Insertion parameterization $\phi$ into a Node Insertion Model with parameters $\phi_1$ and a Halting Model with parameters $\phi_2$.

\subsection{The FLAGG Framework}\label{sec:method/modular}
The factorization trick of Equation~\eqref{eq:broken_remv} is central in FLAGG, highlighting the two irreconcilable and highly correlated components of graph generation: selecting the size and building the topology. To our knowledge, a model that merges them end-to-end doesn't exist. For example, one-shot models first sample the size from the empirical distribution and then build the structure for the selected size. On the other hand, FLAGG breaks down generation into a chain of steps, interleaving the insertion of new nodes with the generation of the topology. In particular, FLAGG's Insertion Model $p_{\phi}(r_t|\gG_t)$ and the Filler Model $p_\theta(\gW_t|r_t,\gG_t)$, are implemented as separate modules, with potentially separate training. For instance, the Filler Model can be implemented by a one-shot model, which also generates the connectivity between graphs. The Insertion Model can be broken down into a halting component and a block size predictive component, as we do in this paper. The whole FLAGG model can be formulated as follows:
\begin{equation}
    \begin{aligned}
    p_{\theta,\phi}(\gG) &=
    \sum_{T=1}^{\infty } \sum_{\gG_{T:0} \in \mathcal{R}(\gG,T)} \underbrace{p_\phi(r_0=0|\gG_{0})}_{\text{halt}} p_\theta(\gG_{T}) \prod_{t=1}^{T} \underbrace{p_\theta(\gW_t|r_t,\gG_t)}_{\text{fill}} \underbrace{p_\phi(r_t|\gG_{t})}_{\text{insert}}.
    \end{aligned}
    \label{eq:ifh}
\end{equation}
The modularity of FLAGG allows for different formulations of each part to work in various settings. These inductive biases are first injected in the removal process, which is designed to act on graphs of different topologies. Fig.~\ref{fig:ifh} shows an overview of FLAGG.

The model in Equation~\eqref{eq:ifh} is trained following Algorithm~\ref{alg:training}. Given an example graph $\gG$, we first sample an entire removal sequence $\gG_{0:T}$. The models parameters $\phi,\theta$ can be optimized by minimizing the variational upper bound $L_\text{vub}(\phi_1,\theta)$, defined in Equation~\eqref{eq:ifh_loss}. We add a binary cross-entropy term $L_\text{halt}$ to learn the halting signal, with parameters $\phi_2$. Sampling from the FLAGG model~\eqref{eq:ifh} is expressed in Algorithm~\ref{alg:sampling} and is a loop of node insertions, connecting the new nodes and choosing to halt the loop.

\input{main-paper/algo-train}
\input{main-paper/algo-sample}

%% file: main-paper/algo-train.tex
\begin{algorithm}[t]
\caption{Training}\label{alg:training}
\begin{algorithmic}[1]
\Repeat
   \State $\gG_0\sim q(\gG)$
   \State $t \gets 1$
   \While {$\gG_{t-1} \neq \varnothing$}
       \State $\gG_{t} \sim q(\gG_{t}|\gG_{t-1})$ \Comment{remove nodes}
       \State $r_t \gets n_{t-1} - n_t$ \Comment{get true number of nodes}
       \State $\gW_{t} \gets \operatorname{split}(\gG_{t-1}, \verts_t)$ \Comment{get true nodes and edges}
       \State $h_{t} \gets \mathbf{1}_{t=1}$ \Comment{get true halting signal 0/1}
       \State $L_{\text{ins},t}(\phi_1) \gets \KL \big(q(r_t|\gG_t,\gG_0) \Vert p_{\phi_1}(r_t|\gG_t)\big)$
       \State $L_{\text{fill},t}(\theta) \gets \KL \big(q(\gW_t|r_t,\gG_t,\gG_0) \Vert p_{\theta}(\gW_t|r_t,\gG_t)\big)$
       \State $L_{\text{halt},t}(\phi_2) \gets L_{\text{halt}}(h_{t}, \lambda_{\phi_2}(\gG_{t-1}))$
   \EndWhile
   \State Perform gradient descent step on \vspace{-6pt}
   \begin{equation*}
   \vspace{-6pt}
   \frac{1}{T}\sum_{t=1}^T (L_{\text{ins},t}(\phi_1) + L_{\text{fill},t}(\theta) + L_{\text{halt},t}(\phi_2))
   \end{equation*}
\Until {converged}
\end{algorithmic}
\end{algorithm}

%% file: main-paper/algo-sample.tex
\begin{algorithm}[t]
\caption{Sampling}\label{alg:sampling}
\begin{algorithmic}[1]
    \State $\gG \gets \varnothing$ \Comment{start from the empty graph}
    \Repeat
        \State $r \sim p_{\phi_1}(r|\gG)$ \Comment{sample how many nodes to add}
        \State $\gW \sim p_{\theta}(\gW|r,\gG)$ \Comment{sample new nodes and edges}
        \State $\gG \gets \operatorname{merge}(\gG, \gW)$
        \State $h \sim \lambda_{\phi_2} (\gG)$ \Comment{sample halting signal}
    \Until {$h=1$}
    \State return $\gG$
\end{algorithmic}
\end{algorithm}

%% file: main-paper/implement.tex
\section{Instances of FLAGG}\label{sec:inst}

In this section, we outline different ways to implement FLAGG. First, we start with existing models in the literature, showing that they also fit into the FLAGG formulation. Then, we describe the adaptation technique for one-shot models to be used as Filler Models, and how the different instance of FLAGG affects memory usage and sampling time. Finally, we introduce two additional variants of FLAGG: one for generating large graphs, and one for sampling graph properties of interest.

\subsection{Specializing FLAGG to Models in the Literature}\label{sec:inst/spec}

One of the aims of FLAGG is to bridge the gap between one-shot and autoregressive models. In this section, we describe how to implement these models with the three blocks of \textit{insertion}, \textit{filling}, and \textit{halting}.

\subsubsection{One-Shot Models}
For one-shot models~\citep{zang2020moflow,liu2021graphebm,martinkus2022spectre,vignac2023digress,jo2022gdss,huang2023cdgs}, all nodes are inserted in one step:
\begin{equation}
    p_{\theta,\phi}(\gG) = p_{\theta}(\gG|n)p_{\phi}(n).
\end{equation}
\begin{itemize}[align=left,left=0pt]
\item[\textbf{Insertion:}] The total number of nodes is sampled from a categorical distribution given by the empirical histogram of graph sizes computed from the training data set.
\item[\textbf{Filling:}] The filler model is the one-shot model itself, given the number of nodes to generate.
\item[\textbf{Halting:}] The model always halts in one step.
\end{itemize}

\subsubsection{1-Node Sequential Models}
These autoregressive models generate one node at a time~\citep{shi2020graphaf,luo2021graphdf,kong2023grapharm}, and exactly take $n$ steps to generate a graph:
\begin{equation}
    \begin{aligned}
    p_{\theta,\phi}(\gG) &= \sum_{\gG_{n:0} \in \mathcal{R}(\gG,n)} \lambda_\phi(\gG_{0}) p_\theta(\gG_n) \prod_{t=1}^{n}(1-\lambda_\phi(\gG_{t})) p_\theta(\gW_t|r_t,\gG_t).
    \end{aligned}
\end{equation}
\begin{itemize}[align=left,left=0pt]
\item[\textbf{Insertion:}] One node is deterministically inserted at each step.
\item[\textbf{Filling:}] Typically, the model samples the node label and an adjacency vector connecting the new node to the current graph. Sometimes, also edges are generated autoregressively~\citep{shi2020graphaf,luo2021graphdf}.
\item[\textbf{Halting:}] This model can be implemented in several ways. For example, in~\citet{you2018graphrnn}, an End-Of-Sequence (EOS) token is sampled to end generation; in~\citet{shi2020graphaf} the number of nodes is fixed at the start, sampled from the empirical distribution; in~\citet{luo2021graphdf} generation stops when a threshold is reached, or if the model does not link the new node to the previous subgraph; \citet{han2023graph_halt} trains a neural network to predict the halting signal from the adjacency matrix.
\end{itemize}

\subsubsection{Hybrid Models}
Other models have tried to fill the gap in autoregressive generation. One of the most well-known is GRAN~\citep{liao2019gran}, which samples the total number of nodes from the empirical distribution, and generates blocks of a fixed size until the target amount is hit. Then, the overflown nodes are removed, leaving the final graph. The block size is fixed as a hyperparameter, allowing the change in the level of sequentiality. Still, this is a restrictive formulation of block generation, as the Insertion Model of GRAN deterministically chooses one block size. FLAGG generalizes these models by allowing different insertion policies, with different block sizes. Incidentally, FLAGG solves the problem of GRAN eliminating nodes at the end, as we may consider only node-adding operations of different sizes.

\subsection{Adapting One-Shot Models for FLAGG}\label{sec:inst/adapting}

As anticipated in Section~\ref{sec:method/rev_proc}, the Filler Model can be implemented through a one-shot model. We recall that the aim is to generate a new block $\gW_{t}=(\gG_t^B,\edges_t^{AB},\edges_t^{BA})$ with the new subgraph $\gG_t^B$ of $r_t$ nodes and the intermediate edges $\edges_t^{AB},\edges_t^{BA}$, and merge it to the previous graph $\gG_t^A$ with $n_t$ nodes (Definition~\ref{def:merge_op}). When the graphs are undirected, we can just consider $\edges_t^{AB}$. Still, a one-shot model only generates a single graph $\gG$, so it must be adapted to also generate the links $\edges_t^{AB},\edges_t^{BA}$. This can be done by first encoding the topological structure of $\gG_{t}^A$ into its nodes using a Graph Neural Network such as GraphConv~\citep{morris2019graphconv}, or RGCN~\citep{schlichtkrull2018rgcn} for labeled data. These encodings are given as input to the architecture of the one-shot model to propagate information from the intermediate graph to the new block. The output of the one-shot model includes the new node labels (if any), the $n_t\times r_t$ rectangular adjacency matrix for intermediate edges, and the $r_t\times r_t$ square adjacency matrix for the new block. After generation, the two partitions are merged and transformed into a sparsely represented partial graph for the next step. This is particularly useful when generating large graphs, as the adjacency matrix can be very sparse. Differently from~\citet{cognolato2024ifh}, we let the architecture update the node representations of the partial graph in its neural network layers, having a back-and-forth of information. This is crucial to account for the evolution of new edges during the generative process, particularly in diffusion-based one-shot models.
An important implication of this adaptation technique is that any one-shot model can be easily extended to be autoregressive.

\subsection{Complexity Considerations}\label{sec:inst/complexity}

Many factors determine the sampling complexity of an instance of FLAGG. Looking at Algorithm~\ref{alg:sampling}, most time is spent generating the new nodes and edges with the Filler Model. For example, in the case of Diffusion Models, which are well known to be time-consuming, having many calls can lead to long sampling times. Parallel hardware like GPUs can accelerate a single sampling call, but repeated calls cannot be parallelized. On the other hand, memory consumption scales quadratically with the number of nodes, which hinders using very large blocks, although the sparse representation helps. The flexibility of FLAGG allows trading off memory for compute time and vice versa.
On the training side (Algorithm~\ref{alg:training}), \textit{batching} multiple examples together can create memory bottlenecks, so a removal process should be carefully designed to avoid this, as is done with the Categorical removal process. As explained in Section~\ref{sec:method/remv_procs}, having a high variability and maximum block size can lead to high memory consumption, as smaller blocks will tend to have a lot of padding. See \citet{cognolato2024ifh} for an empirical evaluation of these effects.

\subsection{Generating Large Graphs}\label{sec:inst/large_graphs}
Thanks to the sparsity of FLAGG, the model can be used to generate very large graphs. Still, memory usage can be further reduced. One way during training is to avoid sampling the entire removal sequences, but sample a subset of time steps, leveraging the closed form of marginal $t$-steps distributions. Another less trivial modification is implementing the node selection mechanism of EDGE~\citep{chen2023efficient}, a one-shot Graph Diffusion Model. In our case, we sample a mask over nodes to select a subset for generation at each step. In this case, the Filler Model will generate the connectivity only to this subset. We test this variant in the Experiments Section~\ref{sec:experim/large_results}.

\subsection{Sequential Generation with Properties}\label{sec:inst/properties}
The factorization trick in Equation~\ref{eq:broken_ins} showed us that a generative step can be broken down into a node insertion operation and a connectivity filling operation. We experiment with the addition of new properties to sample in the chain.
In general, let $P_{1:k}=(P_i)_{i=1}^k$ be a sequence of k properties to sample from an intermediate graph. The filler step will be conditioned on the values of these properties. Then, the FLAGG reverse process can be factorized into the chain:
\begin{equation}
    \begin{aligned}
        p_{\phi,\theta}(\gG_{t-1}|\gG_t) &= p_{\theta}(\gG_{t-1}|P_{1:k}(\gG_{t-1}),\gG_t) p_\phi(P_{1:k}(\gG_{t-1})|\gG_t)\\
        &= p_{\theta}(\gG_{t-1}|P_{1:k}(\gG_{t-1}),\gG_t) \prod_{i=1}^k p_{\phi_i}(P_i(\gG_{t-1})|P_{1:i-1}(\gG_{t-1}),\gG_t),
    \end{aligned}
\end{equation}
where each $p_{\phi_i}$ is a model predicting the property $P_i$ of the new graph, given the already predicted properties.
We already showed the case of the \textit{number of nodes to insert} property, which we name here as $P_1$. For a simpler implementation of this variant of FLAGG, we fix the number of nodes to insert to 1. 
Intending to improve the match with data graphs concerning degree and clustering coefficient distributions, we include the following properties. First, we define $P_2$ as the \textit{current} degree of the newly inserted node. To enforce this constraint on the edges of the node, we add exactly the number of predicted edges in an autoregressive way. For deciding where to place a new edge, the next model predicts $P_3$, a set of preference weights over the remaining nodes, proportional to their degrees. The Barabasi-Albert~\citep{albert2002statistical} models inspire this mechanism, as it incorporates two key properties of real-world networks, i.e., growth and preferential attachment. Growth happens by inserting new nodes, with each new node having an initial degree fixed to $m$. Preferential attachment is ensured by making the probability of connecting to a node proportional to its degree: the larger, the more likely it is to connect. These two principles can capture the power-law degree distribution of networks like the World Wide Web, social networks, and biological networks. Another ingredient is to carefully choose the order of removals of nodes, e.g., an exponential distribution proportional to the degree of nodes. This means we first remove nodes with low degrees and so on to the hubs. From the generation perspective, the model first generates the hubs and then the outer nodes.
Training a model with $P_2$ and $P_3$, and using this order, allows this variant of FLAGG to generate networks following the Barabasi-Albert model.
Finally, to capture local structures we also sample preference weights on the distance from the previously linked nodes, which we call $P_4$. This property is useful when neighbors of a connected node should also be connected, resulting in a better match of node clustering coefficients with the data set. We test this approach in the Experiments Section~\ref{sec:experim/generic_results}.

%% file: main-paper/experiments.tex
\section{Experiments}\label{sec:experim}

In this section, we present the experimental setup and results of FLAGG. We first describe the data sets used for evaluation, the metrics, the baselines considered, and our proposed models. Finally, we present the results and discuss their implications.

\subsection{Data Sets}\label{sec:experim/datasets}

We evaluate FLAGG against the preliminary work~\citet{cognolato2024ifh} and the state of the art, following the same methodology of~\citet{cognolato2024ifh}. Due to its challenges, assessing the quality of generated graphs has been a subject of studies in the literature~\citep{thompson2022evaluation}. In the graph generative task, one would like samples close in probability to those seen in the data set, but this distance is difficult to define unambiguously. The current metrics are based on the Maximum Mean Discrepancy (MMD) between two sets of graphs, computing pairwise kernel functions $k(\cdot,\cdot)$ on features extracted from graphs using a feature extractor $f$. MMD is defined as:
\begin{equation}
    \begin{aligned}
        \text{MMD}^2(p,q) &= \mathbb{E}_{\gG,\gG'\sim p}[k_f(\gG,\gG')] + \mathbb{E}_{\gG,\gG'\sim q}[k_f(\gG,\gG')] - 2\mathbb{E}_{\gG\sim p,\gG'\sim q}[k_f(\gG,\gG')],
    \end{aligned}    
\end{equation}
which can be empirically computed by replacing the expectations with sample averages. The kernel function $k$ that is most often applied is the Radial Basis Function (RBF) kernel, defined as $k(x,x')=\exp(-\gamma\|x-x'\|^2)$. The feature extractor $f$ can be a Neural Network or a function of the graph's adjacency matrix, like the Laplacian Spectrum or the Degree distribution. For this reason, works in this field usually report several metrics, each carrying a different aspect.

\subsubsection{Molecular Data Sets}\label{sec:experim/datasets/molecular}
Molecule generation has been a long-standing problem in the field of graph generation. Molecules present a rich structure, with nodes representing atoms and edges representing bonds. In this manuscript we focus on two popular molecular data sets: QM9~\citep{ramakrishnan2014qm9}, a data set of 133K high-quality small molecules (from 1 to 9 heavy atoms per molecule, with 4 heavy atom types), and ZINC250k~\citep{irwin2012zinc} with 250K larger and less filtered molecules (from 6 to 38 heavy atoms per molecule, with 9 heavy atom types). As is done in other works~\citep{cognolato2024ifh,zang2020moflow,vignac2023digress,shi2020graphaf}, we kekulize the molecules, replacing aromatic bonds with single and double bonds, and remove the hydrogen atoms, using the chemistry library RDKit~\citep{rdkit}. Hydrogen atoms can be easily added back on unpaired electrons of heavy atoms. For both data sets, we generate 10K molecules, and evaluate the following metrics against the respective test sets, and a 10\% of the training set for validation. To measure the quality of sampled molecules we first compute the fraction of \begin{enumerate*}[label=(\roman*)]
    \item valid molecules, which can be reconstructed by RDKit;
    \item unique molecules, that is, the fraction of molecules that are not duplicates;
    \item novel molecules, which are not present in the training set.
\end{enumerate*}
To compute uniqueness and novelty, we use the canonical SMILES representation, a unique string representation for molecules.
We compute the Fréchet ChemNet Distance (FCD) and Neighborhood Subgraph Pairwise Distance Kernel (NSPDK) to evaluate whether the model learned the data set distribution. The two metrics are based on MMD, with FCD getting features from the penultimate layer of ChemNet, a drug activity predictive neural network, and NSPDK accounting for topological features in neighborhood subgraphs. These metrics are computed against the test set.

\subsubsection{Generic Graphs Data Sets}\label{sec:experim/datasets/generic}
We evaluate on graph distributions presenting different topological properties: Community-small~\citep{you2018graphrnn}, with 100 graphs, each with two equally sized communities, linked by a small number of edges; Ego-small and Ego~\citep{sen2008ego}, with 200 and 757 graphs, each representing a social network with highly connected hubs; Enzymes~\citep{schomburg2004enzymes}, with 563 protein graphs, each representing a protein structure. Preprocessing entails making all graphs undirected and stripped from any pre-existing labeling. In this case, the objective is to learn the topological structure of graphs. We split the train/validation/test sets with the 60/20/20 proportion. Following~\citet{huang2023cdgs,cognolato2024ifh}, we compute the MMD with RBF on the distribution of Degree, Clustering coefficient, Laplacian Spectrum coefficient (Spec.), and random GIN embeddings~\citep{thompson2022evaluation}, where the latter can be thought of as a replacement of FCD for generic graphs. For Community-small and Ego-small we generate 1024 graphs, and for Ego and Enzymes we generate the same number of graphs as their respective test sets.

\subsubsection{Large Graphs Generation}\label{sec:experim/datasets/large_graphs}
To push FLAGG to very large graph generation, we experiment on the Cora citation network~\citep{prithviraj2008cora}, with 2810 nodes in the largest connected component. We train FLAGG to generate Cora in an autoregressive way and test whether the statistics match those of the actual network. The metrics used are the same as those in the generic graph data sets, with the addition of Eccentricity, i.e., the average shortest path length. Following~\citet{davies2023size}, we use the total variation kernel (TV) to compute the MMD of the properties.

\subsection{Evaluating FLAGG}\label{sec:experim/setup}
We implemented FLAGG using PyTorch~\citep{paszke2019pytorch}, PyTorch Lightning~\citep{Falcon_PyTorch_Lightning_2019} and PyTorch Geometric~\citep{torch_geom}. The configurations for our experiments can be found as Hydra~\citep{Yadan2019Hydra} files. We run each experiment on an A40 GPU, with three seeds from 0 to 2, ensuring reproducibility. The source code is available at \url{https://github.com/CognacS/flagg-graphgen}.

\subsubsection{Vanilla FLAGG}
We evaluate the model over the same levels of sequentiality of~\citet{cognolato2024ifh} for comparison. These can be found in the rows of Tables~\ref{tab:qm9_results} and~\ref{tab:zinc_results} for molecular data sets, and in Table~\ref{tab:seq_levels} for generic graphs data sets. We chose not to consider the 1-node sequential and one-shot variants as we found these do not benefit much from the improvements we proposed. We pick DiGress~\citep{vignac2023digress} with the optimal prior as the Filler Model, differently from IFH which used DiGress with the uniform prior. We fix the node ordering as the Breadth First Search (BFS) order to enforce the presence of a single connected component, and use the Categorical Removal of Section~\ref{sec:method/remv_procs}, as it is the most efficient and performant~\citep{cognolato2024ifh}. Finally, we implement the new formulation of the Filler Model proposed in Section~\ref{sec:inst/adapting}, and add new features as input to all neural architectures. These include topological features like the number of cycles of different lengths, and spectral features like eigenvectors and eigenvalues of the Laplacian matrix. The Filler Model is implemented through a Graph Transformer~\citep{dwivedi2020graph_transf}, the Insertion Model as an RGCN~\citep{schlichtkrull2018rgcn}, and the Halting Model as an MLP, only taking global features as input. To overcome the instabilities found in IFH, we integrate techniques such as Exponential Moving Average (EMA) and gradient clipping. To stop training at the right time, we use early stopping. As the stopping criteria, we use the best FCD for molecules and the best GIN MMD for generic graphs, as they offer a comprehensive view of the graph's similarities in the respective domains.
\subsubsection{Sparse FLAGG}
We keep the same components of the vanilla variant, but reduce the number of sampled steps during training to 5, and fix the block sizes to $\{1,4,16,32\}$. Before calling the Filler Model to generate edges, our model samples a mask over nodes, reducing the number of nodes considered by the Filler Model. Additionally, we replace the BFS ordering for the exponential degree distribution ordering introduced in Section~\ref{sec:inst/properties}. From the Barabasi-Albert Model~\citep{albert2002statistical} point of view, such an ordering emulates that which generated large networks like Cora.
\subsubsection{Properties FLAGG}
As anticipated in Section~\ref{sec:inst/properties}, we factorize FLAGG into submodules, each sampling a specific property or preference weight. Each of these submodules is implemented through an RGCN. During linkage, edges are added autoregressively, and the probability is proportional to the weights predicted by the model. In particular, the weights depend on the nodes' degree and the distance from the already-linked nodes. Finally, the weights are summed, and the edge is sampled from a categorical distribution with the Softmax of the weights as probabilities. We found that summing the weights instead of computing the chain rule of probabilities leads to better results. Again, we use the exponential degree ordering for this variant. We evaluate Properties-FLAGG (P-FLAGG) on the generic graphs' data sets, also comparing with vanilla FLAGG.

\begin{table}[t]
    \centering
    \begin{tabular}{ccccc}
    \toprule
        Method & Ego-small & Enzymes & Ego & Comm-small \\
        \midrule
        seq-1 & 1 & 1 & 1 & 1\\
        seq-small & $\{1, 2\}$ & $\{1, 3\}$ & $\{1, 3\}$ & \{1, 2\}\\
        seq-big & $\{1, 2, 8\}$ & $\{1, 2, 8\}$ & $\{1, 4, 16\}$ & \{1, 2, 8\}\\
        one-shot & $n$ & $n$ & $n$ & $n$\\ 
    \bottomrule
    \end{tabular}%
    \caption{Sequentiality levels on generic graphs, i.e., block sizes used.}\label{tab:seq_levels}
\end{table}

% insert here big table with results
\input{main-paper/tables/res_table}

\subsection{Results on Molecular Data Sets}\label{sec:experim/mol_results}
\subsubsection{Baselines}
On molecular data sets, we compare against most of the state-of-the-art models reported in~\citet{huang2023cdgs}. Specifically, we consider the autoregressive models GraphAF~\citep{shi2020graphaf}, GraphDF~\citep{luo2021graphdf}, GraphARM~\citep{kong2023grapharm}, and the one-shot models MoFlow~\citep{zang2020moflow}, DiGress~\citep{vignac2023digress}, CDGS~\citep{huang2023cdgs}. We also compare against IFH~\citep{cognolato2024ifh}, the work we are extending.

\subsubsection{Analysis}

Looking at Tables~\ref{tab:qm9_results} and~\ref{tab:zinc_results}, we observe a substantial improvement of intermediate levels of sequentiality going from IFH to FLAGG, which are now on par with the 1-node sequential variant of IFH. FCD results are the highlights of FLAGG, surpassing all other models in both QM9 and ZINC using large block sizes. This induces us to think that the model is better at incorporating the molecular structure than CDGS, and indeed better than IFH. A reason for the improvement over IFH can be found in the architectural modification introduced in Section~\ref{sec:inst/adapting}, which allows better information flow and makes the task easier to learn. Regarding the remaining metrics, FLAGG is competitive with the other models, with a validity slightly lower in ZINC.

\subsection{Results on Generic Graphs Data Sets}\label{sec:experim/generic_results}
\subsubsection{Baselines}
On generic graphs data sets, we compare vanilla FLAGG and P-FLAGG with the autoregressive models GraphRNN~\citep{you2018graphrnn}, GRAN~\citep{liao2019gran}, and the one-shot models VGAE~\citep{simonovsky2018graphvae}, EDP-GNN~\citep{niu2020edp_gnn}, GDSS~\citep{jo2022gdss}, CDGS~\citep{huang2023cdgs}. Again, we also compare against IFH~\citep{cognolato2024ifh}, as it underperformed by a great margin for all levels of sequentiality.

\subsubsection{Analysis}

Comparing our vanilla FLAGG variants with those of IFH in Table~\ref{tab:generic_graphs_results}, we see a massive improvement in all metrics. Not only we managed to improve all intermediate models, but we also achieved better results than CDGS in the Ego-small and Enzymes data sets. Again, we argue that the improvements introduced in this work are the reason for the better results. Still, we observe that for Community-small and Ego, our vanilla FLAGG lags behind CDGS. In particular, the Degree and Clustering Coefficient metrics are lower, meaning that the model generates a slightly different number of edges and potentially does not fully capture the community structure measured by the clustering coefficients. Then, we extend the comparison with P-FLAGG, which explicitly learns these features. We want to be clear that the model is not \textit{informed} about which property value is the right one, but rather, this decision is made explicit in the factorization and is learned during training. Looking at the two above-mentioned metrics, we observe an improvement over all data sets but Ego-small, where results are already satisfying. This, combined with its low sampling time, makes P-FLAGG a promising direction for future research.

\subsection{Results on Huge Graphs Generation}\label{sec:experim/large_results}
\subsubsection{Baselines}
We compare large-scale generation of Cora against HiGGs~\citep{davies2023size}. For this purpose, we defined a sparse version of FLAGG, which is more memory efficient and can generate graphs of this size.

\subsubsection{Analysis}
Results on this challenging benchmark, shown in Table~\ref{tab:cora_results}, suggest that FLAGG captures very well the degree distribution of Cora and its spectral information but lags in terms of Clustering Coefficients and Eccentricity. This could be explained by HiGGs's algorithm, which first generates the skeleton of the network, and finally expands the substructures. This might help in capturing the local structure of the graph. On the other hand, first generating the communities, and then linking them in parallel might overshoot the number of edges generated, motivating the worse Degree metric of HiGGs.

\input{main-paper/fig_grids/comm}
\input{main-paper/fig_grids/egosm}
\input{main-paper/fig_grids/enz}
\input{main-paper/fig_grids/ego}
\input{main-paper/fig_grids/qm9}
\input{main-paper/fig_grids/zinc}

\subsection{Qualitative Results}\label{sec:experim/qualitative}

After discussing the quantitative results in the previous section, we briefly evaluate the quality of generated samples from a qualitative perspective. Figures~\ref{fig:comm_samples},~\ref{fig:egosm_samples},~\ref{fig:enz_samples},~\ref{fig:ego_samples},~\ref{fig:qm9_samples}, and~\ref{fig:zinc_samples} show random batches generated by each of our models, together with random samples from the training datasets. Generally, we observe that geometric properties are respected, and molecular structures are valid.

\subsection{Sampling Efficiency}\label{sec:experim/block_size}
\subsubsection{Sampling time}

In Table~\ref{tab:sampling_time} we report the sampling time of all FLAGG variants on all data sets. A clear trend can be observed: sampling time decreases as the block size increases. This reflects the considerations made in Section~\ref{sec:inst/complexity}, as using larger blocks drastically reduces the number of sequential steps, and increases parallelization on GPU. A notable exception is the Properties-FLAGG, which is orders of magnitude faster than other methods, although being fully sequential, i.e., one node and one edge at a time. For instance, in the Ego dataset \texttt{properties} FLAGG's time is roughly 225 times smaller than \texttt{seq-small} FLAGG. This is due to the algorithm being fully autoregressive, avoiding the use of expensive Diffusion Models. We investigate whether we can obtain speed-ups in Diffusion-based models in the next section.

\input{main-paper/tables/sampling_time}

\subsubsection{Compute Time vs. Performance}

\input{main-paper/fig_grids/tradeoff}

We investigate whether a speed-up in Diffusion-based FLAGG variants can be obtained. In particular, we explore the use of a lighter Filler Model, evaluating its sampling time and how it affects the sampling quality. For this purpose, we pick the best performing FLAGG variant on the QM9 molecular data set, i.e., the $\{1,4\}$-blocks model, and we vary the number of denoising steps performed by DiGress. As explained in~\cite{nichol2021improveddiff}, it is possible to reduce the number of denoising steps from $T$ to $K=T/J$ by defining a jump $J$ such that 1 step in the reduced process corresponds to $J$ steps in the original process. We define a set of reduced numbers of steps $K=1,2,5,10,20,50,100,250,500$, where $T=500$. Results in Figure~\ref{fig:perf_vs_time} show that reducing the number of steps by up to a factor of $J=10$ keeps the validity over $99\%$ and the FCD at a very low value, while reducing the sampling time by a factor of $\sim 10$. After this threshold, FCD and validity start to rapidly deteriorate until, at $K=1$ step, validity sits at $55.4\%$ and FCD at $6.788$. This behavior is consistent with what was found in~\cite{nichol2021improveddiff} for image generation with diffusion models. Regarding the relation between $K$ and sampling time, it can be seen that they are almost always directly proportional, with $J$ being the proportionality constant. This is not a perfect relation as the filler model is only a submodule of FLAGG, and fixed costs for insertion and halting are always present.

\subsection{Ablation Studies}
\subsubsection{On the Node Ordering}
\input{main-paper/tables/ablation}

Recalling from Section~\ref{sec:method/remv_procs/order}, the node ordering is fixed by the experimenter as a hyperparameter of the removal process, in our case the BFS ordering. Then, FLAGG learns to reverse the particular ordering used. Thus, we investigate the impact of this particular ordering. For this purpose, we train two additional variants of the Vanilla FLAGG model on the QM9 dataset with the best performing set of blocks (i.e., $\{1,4\}$). The first variant uses the Depth First Search (DFS) ordering, which has also been investigated in~\cite{liao2019gran}, and the second variant uses a simple random ordering. Results comparing BFS, DFS and random are shown in Table~\ref{tab:ablation}. With respect to molecular validity, all orderings reach $>99\%$, so the true comparison must be made on the remaining metrics. BFS achieves the best value of FCD and NSPDK, suggesting that it was better suited for learning the molecular structure of QM9. Still, DFS comes close to BFS, while the random ordering is worse, especially in terms of FCD, where it is higher by $75\%$.

\subsubsection{On the Halting Model}
\input{main-paper/tables/halting_base}

The halting model is a critical part of the FLAGG framework. Thus, we evaluate the same framework using a static insertion process, sampling the total number of nodes from the empirical distribution of the training set. Consequently, the block sizes to be inserted are sampled from the removal process, as defined in Sec.~\ref{sec:method/remv_procs/cat}. Results are shown in Table~\ref{tab:halting_baseline}, using our trained checkpoints on the Enzymes dataset, and replacing the learned components with empirical ones as described above. An immediate observation is that the learned model dominates on the local metrics like Degree and Clustering Coefficients, while the empirical one performs better on global metrics, i.e., the spectrum and the GIN embeddings. This suggests that trained halting and insertion models may allow a better replication of local structures from the dataset, but, if not perfectly mirroring the empirical distribution, they might lead to a shift in the graph distribution, resulting in different global properties. We check this shift in graph sizes in the next section.

\subsection{Analysis of the Learned Insertion Policy}

We investigate the insertion policy of FLAGG by studying the learned behaviors of its three modules:
\begin{enumerate*}[label=(\roman*)]
    \item the Halting model's predicted distribution over the number of nodes, and how it compares to the empirical distribution of the training set;
    \item the preferred block sizes of the Insertion model, depending on time;
    \item the Filler model's learned node ordering.
\end{enumerate*}
We ground the analysis on the Enzymes dataset, as it presents medium-to-high complexity relative to the presented data sets. It has a non-trivial empirical distribution of the number of nodes, and graphs have both long chains and cycles. We generate a batch of $N$ graphs $\mathcal{B}=\{\gG^1,\ldots,\gG^N\}$, keeping track of intermediate graphs $(\gG^i_t)_{t=0}^T$, where we reverse the ordering of time for clarity (i.e., $\gG^i_0=\emptyset$ and $\gG^i_T=\gG^i$), and we refer to it in the following sections.

\subsubsection{On Halting and Graph Sizes}

\input{main-paper/fig_grids/halt_grid}

First, we check whether the halting model, together with the insertion model, can reconstruct the empirical distribution of the number of nodes. Relevant plots can be found in Figure~\ref{fig:halt_grid}. We compare two different distributions computed from $\mathcal{B}$: the \textit{Generated} node-count frequencies $h_\text{gen}(n)=|\{i: |\verts^i|=n\}|/N$, and the distribution computed from the halting \textit{Prior} distribution:
\begin{align}
    p^i_\text{prior}(n) &= \sum_{t:\ |\verts^i_t|=n} \lambda_{\phi_2}(\gG_{t}^i)\prod_{s=1}^{t-1} (1-\lambda_{\phi_2}(\gG_{s}^i)), \\
    h_\text{prior}(n) &= \frac{1}{N}\sum_{i=1}^N p^i_\text{prior}(n),
\end{align}
where $\lambda_{\phi_2}(\gG)$ are the halting probabilities produced by the model to create the batch $\mathcal{B}$. From Figure~\ref{fig:halt_grid}, we can see that the support of both the Prior and Generated histograms matches that of the Dataset, although the probability mass leans towards lower values, meaning the model is sometimes undershooting. An interesting fact that can be observed is how the use of blocks shapes the histograms: in the $\{1,3\}$ case, both the Prior and Generated histograms present peaks with an interval of 3, which is the most frequent block size judging from the trajectories. The same happens in the $\{1,2,8\}$ case, with an interval of $2$. This suggests that the insertion policy is overconfident on block predictions, preferring more frequent block sizes. Still, we don't see signs of overshooting, i.e., graphs larger than the dataset's limit. On the bottom row, we show the trajectories created by the graphs in batch $\mathcal{B}$ with respect to the growth in size over time. For the case of $\{1,3\}$ we can in fact see that graph sizes rapidly evolve by picking the biggest block size, with the halting model contributing to the shape of the histogram by stopping at convenient times. In the $\{1,2,8\}$ regime, block sizes are more heterogeneous, and we even see that getting close to the upper bound of the number of nodes, the model starts to prefer smaller blocks sizes, and finally halting generation. We continue the analysis of block sizes and node ordering in the next sections.

\subsubsection{On Block Sizes and Node Ordering}

\input{main-paper/fig_grids/blocks_grid}

We now turn to the policy on block sizes and node ordering. We refer the reader to Figure~\ref{fig:blocks_grid}. The plots in the top row confirm the hypotheses of the previous section, showing that the $\{1,3\}$ model prefers the block size of 3, while the $\{1,2,8\}$ model varies among the three. From this plot, we can also observe the behavior of picking only the $2$ block at the end of generation by the $\{1,2,8\}$ model. The graphs in Figure~\ref{fig:blocks_grid} show the order in which nodes and edges are generated at each step. We can see that in most graphs, the model mimics the BFS ordering, for example in long chains, where either they are generated from one end to the other, or they are starting from the middle and growing in both directions. Still, we observe that in some cases, particularly when having cycles, blocks forward in time are connected with older blocks. We argue that this is caused by the model not having memory on the order in which nodes were generated.

\subsection{Discussion}\label{sec:experim/discussion}
From the experimental results, we can deduce that FLAGG with nested DiGress is a strong model, managing to generate high-quality graphs with an autoregressive pattern. This can be useful for learning the distribution over nodes in a conditional generation setup, where we can't use the empirical distribution. FLAGG may also be applied to expand already existing graphs.
The improvement achieved by FLAGG, relative to IFH~\citep{cognolato2024ifh}, can be attributed to the architectural modifications, the new topological and spectral features, and the use of techniques like EMA and gradient clipping, as clearly shown in Table~\ref{tab:general}. Results also suggest that not having built-in permutation invariance in the model did not hinder its sample quality. Furthermore, we notice that changing the block sizes does not significantly impact the model's performance. This suggests that, differently from the points raised in~\citet{cognolato2024ifh}, and in the setup of the present work, FLAGG is not particularly sensitive to the choice of the sequentiality level. We advise more research to pinpoint whether this holds for other removal processes, orderings, and filler models.

Another strength of FLAGG is the flexibility of its design. In~\citet{cognolato2024ifh} we showed that increasing the number of steps leads to lower memory consumption but higher time complexity. In this work we showed that the implementation can also be tailored around a particular task, or to boost the similarity with the data set for some property. For example, P-FLAGG improved its adherence to the degree and clustering distribution of the data graphs, thanks to explicitly learning them. Additionally, P-FLAGG proved to be a light-weight model, reducing sampling time by 2 orders of magnitude. Another advantage of FLAGG is the capability to generate very large graphs, as shown with the Cora data set. On this front, we still see the ground for improvement, as the Clustering metric was worse than HiGGs, which can better capture local information.

For future research, we would like to explore new removal processes. For instance, knowing that the data set consists of communities can be exploited in the generative process. This could be built into the removal process by interleaving the ordering of removals with the block sizes. Another direction is to increase the number of topological properties that FLAGG can incorporate, with the hope of having a better interaction between the symbolic and neural aspects of the model.
A current limitation of FLAGG is that the removal process parameters are fixed by the experimenter. One way to learn them is through hyperparameter search, which may prove expensive. Thus, a direct extension of the insertion process would have the model learn an optimal ordering of nodes and block sizes directly from the data, avoiding the need for manual tuning. This extension requires major efforts and additional theoretical background on the specifics, so we leave it as future work.

%% file: main-paper/tables/res_table.tex
\begin{sidewaystable*}[p]
    \begin{minipage}[t]{0.39\textwidth}
    \subfloat[Performance results on the QM9 data set]{%
    \resizebox{\textwidth}{!}{%
    \begin{tabular}{*{7}{c}}
        \toprule
        \multicolumn{2}{c}{Method} & Valid (\%)$\uparrow$ & NSPDK$\downarrow$ & FCD$\downarrow$ & Unique (\%)$\uparrow$ & Novel (\%)$\uparrow$\\
        \midrule
        \multicolumn{2}{c}{Metrics on Training Set} & --- & $1.36\text{e-}4$ & $0.057$ & --- & ---\\
        \midrule
        \multirow{3}{*}{ Autoreg. }
        & GraphAF & $74.43$ & $0.021$ & $5.625$ & $88.64$ & $86.59$\\
        & GraphDF & $93.88$ & $0.064$ & $10.928$ & $98.58$ & $98.54$\\
        & GraphARM & $90.25$ & $0.002$ & $1.220$ & $95.62$ & $70.39$\\
        \midrule
        \multirow{3}{*}{ One-shot }
        & MoFlow & $91.36$ & $0.017$ & $4.467$ & $98.65$ & $94.72$\\
        & DiGress & $99.00$ & $5.00\text{e-}4$ & $0.360$ & $96.66$ & $33.40$\\
        & CDGS & $\underline{99.68}$ & $\underline{3.08\text{e-}4}$ & $\underline{0.200}$ & $96.83$ & $69.62$\\
        \midrule
        \multirow{4}{*}{ IFH }
        & seq-1    & $\mathbf{99.92}$ & $\mathbf{2.99\text{e-}4}$ & $0.902$ & $96.63$ & $88.33$\\
        & \{1, 2\} & $94.34$ & $4.19\text{e-}4$ & $0.904$ & $97.08$ & $89.11$\\
        & \{1, 4\} & $92.51$ & $7.53\text{e-}4$ & $0.995$ & $97.72$ & $92.16$\\
        & one-shot & $95.31$ & $0.002$  & $1.512$ & $96.93$ & $94.65$\\
        \midrule
        \multirow{2}{*}{ FLAGG }
        & \{1, 2\} & $99.49_{0.07}$ & $4.92\text{e-}4_{1.35\text{e-}4}$ & $0.212_{0.040}$ & $97.02_{0.11}$ & $76.92_{0.21}$\\
        & \{1, 4\} & $\underline{99.69}_{0.17}$ & $\underline{3.05\text{e-}4}_{0.36\text{e-}4}$ & $\mathbf{0.144}_{0.010}$ & $96.96_{0.37}$ & $76.04_{0.91}$\\
        \bottomrule
    \end{tabular}
    }\label{tab:qm9_results}
    }
    \end{minipage}
    \begin{minipage}[t]{0.38\textwidth}
    \subfloat[Performance results on the ZINC250K data set]{%
        \resizebox{\textwidth}{!}{%
            \begin{tabular}{*{7}{c}}
                \toprule
                \multicolumn{2}{c}{Method} & Valid (\%)$\uparrow$ & NSPDK$\downarrow$ & FCD$\downarrow$ & Unique (\%)$\uparrow$ & Novel (\%)$\uparrow$\\
                \midrule
                \multicolumn{2}{c}{Metrics on Training Set} & --- & $5.91\text{e-}5$ & $0.985$ & --- & ---\\
                \midrule
                \multirow{3}{*}{ Autoreg. }
                & GraphAF & $68.47$ & $0.044$ & $16.023$ & $98.64$ & $99.99$\\
                & GraphDF & $90.61$ & $0.177$ & $33.546$ & $99.63$ & $100.00$\\
                & GraphARM & $88.23$ & $0.055$ & $16.260$ & $99.46$ & $100.00$\\
                \midrule
                \multirow{3}{*}{ One-shot }
                & MoFlow & $63.11$ & $0.046$ & $20.931$ & $99.99$ & $100.00$\\
                & DiGress & $91.02$ & $0.082$ & $23.06$ & $81.23$ & $100.00$\\
                & CDGS & $\underline{98.13}$ & $\mathbf{7.03\text{e-}4}$ & $\underline{2.069}$ & $99.99$ & $99.99$\\
                \midrule
                \multirow{4}{*}{ IFH }
                & seq-1 & $\mathbf{98.56}$ & $0.002$ & $2.387$ & $99.87$ & $99.89$\\
                & \{1, 3\} & $80.59$ & $0.004$ & $3.312$ & $99.98$ & $99.95$\\
                & \{1, 4, 8\} & $65.68$ & $0.015$ & $9.229$ & $99.94$ & $100.00$\\
                & one-shot & $60.48$ & $0.033$ & $15.174$ & $100.00$ & $100.00$\\
                \midrule
                \multirow{2}{*}{ FLAGG }
                & \{1, 3\} & $97.51_{1.13}$ & $0.002_{4\text{e-}4}$ & $\underline{2.096}_{0.202}$ & $99.97_{0.04}$ & $99.98_{6\text{e-}5}$ \\
                & \{1, 4, 8\} & $97.81_{0.09}$ & $\underline{14.85\text{e-}4}_{3\text{e-}5}$ & $\mathbf{1.810}_{0.030}$ & $100.00_{6\text{e-}5}$ & $99.98_{6\text{e-}5}$ \\
    
                \bottomrule
            \end{tabular}
        }\label{tab:zinc_results}
    }
    \end{minipage}
    \begin{minipage}[t]{0.2\textwidth}
    \subfloat[Performance results on the Cora data set]{%
        \resizebox{\textwidth}{!}{%
            \begin{tabular}{*{7}{c}}
                \toprule
                Method & Deg.$\downarrow$ & Clus.$\downarrow$ & Eccent.$\downarrow$ & Spec.$\downarrow$\\
                \midrule
                HiGGs & $0.254$ & $\mathbf{0.958}$ & $\mathbf{0.273}$ & --\\
                Sparse FLAGG & $\mathbf{0.020}_{0.013}$ & $2.0_{0.0}$ & $0.751_{0.061}$ & $\mathbf{0.025}_{0.008}$\\
                \bottomrule
            \end{tabular}
        }\label{tab:cora_results}%
    }
    \end{minipage}
    
    \vspace{1cm}
    
    \begin{minipage}[t]{\textwidth}
    \subfloat[Performance results on generic graphs data sets]{%
    \resizebox{\textwidth}{!}{%
    \begin{tabular}{*{21}{c}}
        \toprule
        & & \multicolumn{4}{c}{ Community } && \multicolumn{4}{c}{ Ego-small } && \multicolumn{4}{c}{ Enzymes } && \multicolumn{4}{c}{ Ego } \\
        \cline{3-6} \cline{8-11} \cline{13-16} \cline{18-21}
        & & \multicolumn{4}{c}{ $|V|_{\max}=20,\;|E|_{\max}=62$ } && \multicolumn{4}{c}{ $|V|_{\max}=17,\;|E|_{\max}=66$ } && \multicolumn{4}{c}{ $|V|_{\max}=125,\;|E|_{\max}=149$ } && \multicolumn{4}{c}{ $|V|_{\max}=399,\;|E|_{\max}=1071$ }\\
        & & \multicolumn{4}{c}{ $|V|_{\operatorname{avg}}\approx 15,\;|E|_{\operatorname{avg}}\approx 36$ } && \multicolumn{4}{c}{ $|V|_{\operatorname{avg}}\approx 6,\;|E|_{\operatorname{avg}}\approx 9$ } && \multicolumn{4}{c}{ $|V|_{\operatorname{avg}}\approx 33,\;|E|_{\operatorname{avg}}\approx 63$ } && \multicolumn{4}{c}{ $|V|_{\operatorname{avg}}\approx 145,\;|E|_{\operatorname{avg}}\approx 335$ }\\
        \cline{3-6} \cline{8-11} \cline{13-16} \cline{18-21}
        \multicolumn{2}{c}{Method} & Deg.$\downarrow$ & Clus.$\downarrow$ & Spec.$\downarrow$ & GIN$\downarrow$ && Deg.$\downarrow$ & Clus.$\downarrow$ & Spec.$\downarrow$ & GIN$\downarrow$ && Deg.$\downarrow$ & Clus.$\downarrow$ & Spec.$\downarrow$ & GIN$\downarrow$ && Deg.$\downarrow$ & Clus.$\downarrow$ & Spec.$\downarrow$ & GIN$\downarrow$\\
        \midrule
        \multicolumn{2}{c}{Metrics on Training Set}  & $0.035$ & $0.067$ & $0.045$ & $0.037$ && $0.025$ & $0.029$ & $0.027$ & $0.016$ && $0.011$ & $0.011$ & $0.011$ & $0.007$ && $0.009$ & $0.009$ & $0.009$ & $0.005$ \\
        \midrule
        \multirow{2}{*}{ A-R }
        & GraphRNN & $0.106$ & $0.115$ & $0.091$ & $0.353$ && $0.155$ & $0.229$ & $0.167$ & $0.472$ && $0.397$ & $0.302$ & $0.260$ & $1.495$ && $0.140$ & $0.755$ & $0.316$ & $1.283$\\
        & GRAN & $0.125$ & $0.164$ & $0.111$ & $0.196$ && $0.096$ & $0.072$ & $0.095$ & $0.106$ && $0.215$ & $0.147$ & $0.034$ & $0.069$ && $0.594$ & $0.425$ & $1.025$ & $0.244$ \\
        \midrule
        \multirow{4}{*}{ O-S }
        & VGAE & $0.391$ & $0.257$ & $0.095$ & $0.360$ && $0.146$ & $0.046$ & $0.249$ & $0.089$ && $0.811$ & $0.514$ & $0.153$ & $0.716$ && $0.873$ & $1.210$ & $0.935$ & $0.520$ \\
        & EDP-GNN & $0.100$ & $0.140$ & $0.085$ & $0.125$ && $0.026$ & $0.032$ & $0.037$ & $\underline{0.031}$ && $0.120$ & $0.644$ & $0.070$ & $0.119$ && $0.553$ & $0.605$ & $0.374$ & $0.295$\\
        & CDGS & $\mathbf{0.052}$ & $\underline{0.080}$ & $0.064$ & $\mathbf{0.062}$ && $\underline{0.025}$ & $0.031$ & $0.033$ & $\mathbf{0.025}$ && $\underline{0.048}$ & $0.070$ & $0.033$ & $\underline{0.024}$ && $\underline{0.036}$ & $0.075$ & $\mathbf{0.026}$ & $\mathbf{0.026}$ \\
        \midrule
        \multirow{4}{*}{ IFH }
        & seq-1 & $0.209$ & $0.189$ & $0.082$ & $0.277$ && $0.069$ & $0.084$ & $0.066$ & $0.046$ && $\underline{0.049}$ & $\underline{0.049}$ & $\mathbf{0.026}$ & $0.088$ && $0.303$ & $0.643$ & $0.311$ & $0.352$ \\
        & seq-small & $0.177$ & $0.167$ & $0.082$ & $0.203$ && $0.031$ & $0.041$ & $0.040$ & $0.043$ && $0.252$ & $0.237$ & $0.077$ & $0.404$ && $0.435$ & $0.898$ & $0.162$ & $0.403$ \\
        & seq-big & $0.141$ & $0.173$ & $0.089$ & $0.262$ && $0.027$ & $0.042$ & $\underline{0.029}$ &  $0.043$ && $0.441$ & $0.470$ & $0.196$ & $0.698$ && $0.276$ & $0.992$ & $0.190$ & $0.479$\\
        & oneshot & $0.125$ & $0.187$ & $0.081$ & $0.138$ && $0.045$ & $0.065$ & $0.048$ & $0.048$ && $0.264$ & $0.436$ & $0.050$ & $0.180$ && $0.372$ & $0.695$ & $0.458$ & $0.528$ \\
        \midrule
        \multirow{2}{*}{ FLAGG }
        & seq-small & $0.091_{0.004}$ & $0.108_{0.009}$ & $\underline{0.051}_{0.002}$ & $0.126_{0.006}$ && $\mathbf{0.023}_{0.002}$ & $\mathbf{0.015}_{0.001}$ & $\mathbf{0.023}_{0.002}$ & $\underline{0.031}_{0.004}$ && $\underline{0.048}_{0.012}$ & $\mathbf{0.035}_{0.007}$ & $\mathbf{0.026}_{0.004}$ & $\mathbf{0.022}_{0.004}$ && $0.060_{0.018}$ & $\underline{0.061}_{0.011}$ & $0.099_{0.052}$ & $0.075_{0.009}$ \\
        & seq-big & $0.092_{0.010}$ & $0.085_{0.018}$ & $0.056_{0.001}$ & $0.113_{0.016}$ && $0.027_{0.002}$ & $\mathbf{0.015}_{0.002}$ & $\underline{0.028}_{0.002}$ & $0.033_{0.001}$ && $0.071_{0.014}$ & $0.061_{0.022}$ & $0.042_{0.011}$ & $0.054_{0.016}$ && $0.124_{0.027}$ & $0.177_{0.045}$ & $0.117_{0.047}$ & $0.051_{0.008}$ \\
        \midrule
        \multicolumn{2}{c}{ FLAGG with properties }
        & $\underline{0.073}_{0.002}$ & $\mathbf{0.069}_{0.006}$ & $\mathbf{0.049}_{0.003}$ & $\underline{0.094}_{0.010}$ && $0.028_{0.003}$ & $\underline{0.020}_{0.002}$ & $\underline{0.029}_{0.003}$ & $0.034_{0.004}$ && $\mathbf{0.030}_{0.008}$ & $0.067_{0.028}$ & $\underline{0.030}_{0.004}$ & $\mathbf{0.022}_{0.003}$ && $\mathbf{0.033}_{0.001}$ & $\mathbf{0.057}_{0.015}$ & $\underline{0.029}_{0.017}$ & $\underline{0.037}_{0.004}$ \\
        \bottomrule
    \end{tabular}%
    }\label{tab:generic_graphs_results}%
    }
    \end{minipage}
    \caption{Results on the molecule generation task on QM9 (a), ZINC250k (b), Cora (c) and generic graphs (d) averaged over 3 runs. Best results are in bold, and the second best are underlined. Values are reported as $\text{mean}_{\text{std}}$.}\label{tab:general}
\end{sidewaystable*}

%% file: main-paper/fig_grids/comm.tex
\begin{figure}[p]
    \centering
    \begin{subfigure}[t]{0.45\textwidth}
        \centering
        \includegraphics[width=\textwidth]{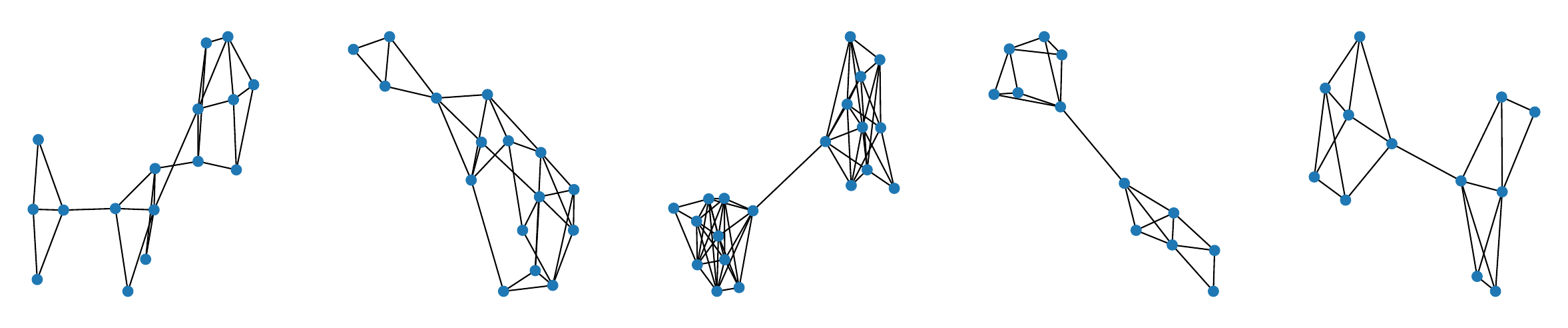}
        \caption{Seq-big FLAGG}
    \end{subfigure}
    \hspace{1cm}
    \begin{subfigure}[t]{0.45\textwidth}
        \centering
        \includegraphics[width=\textwidth]{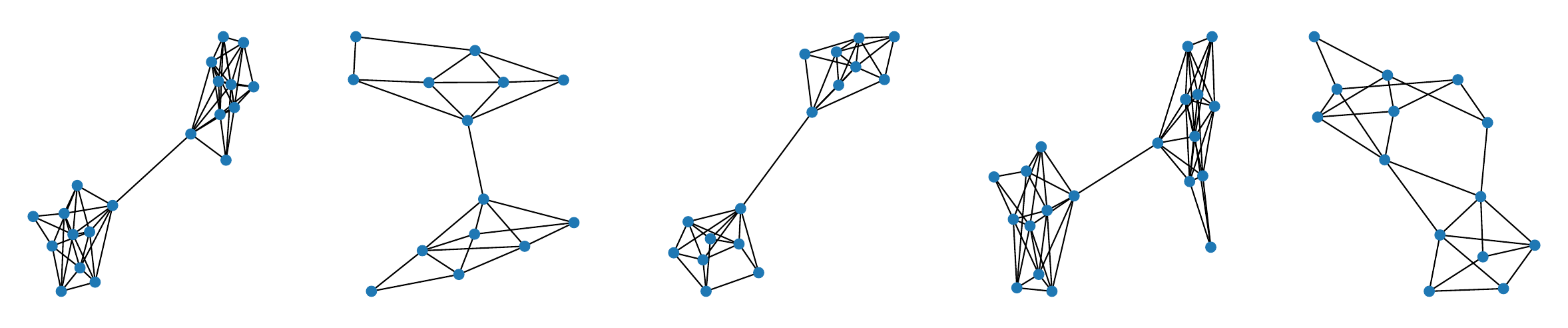}
        \caption{Properties FLAGG}
    \end{subfigure}

    \begin{subfigure}[t]{0.45\textwidth}
        \centering
        \includegraphics[width=\textwidth]{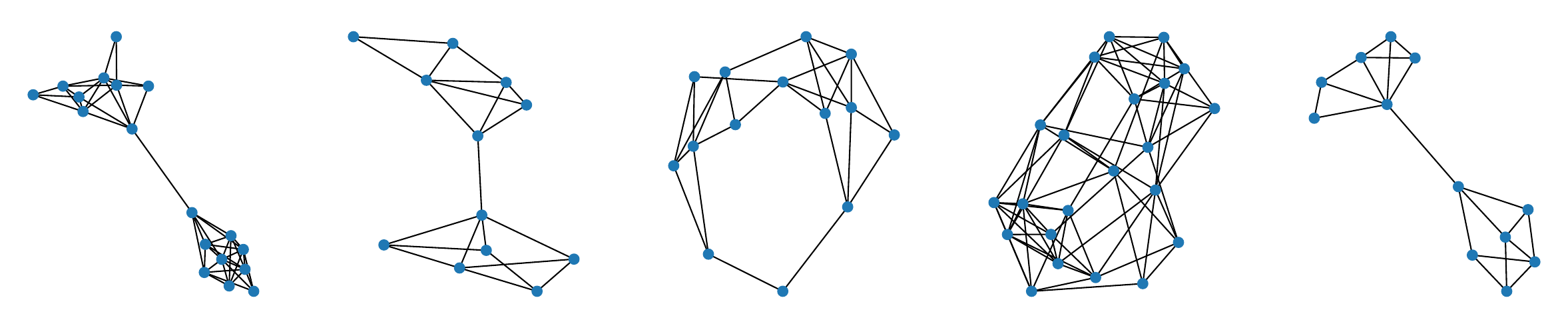}
        \caption{Seq-small FLAGG}
    \end{subfigure}
    \hspace{1cm}
    \begin{subfigure}[t]{0.45\textwidth}
        \centering
        \includegraphics[width=\textwidth]{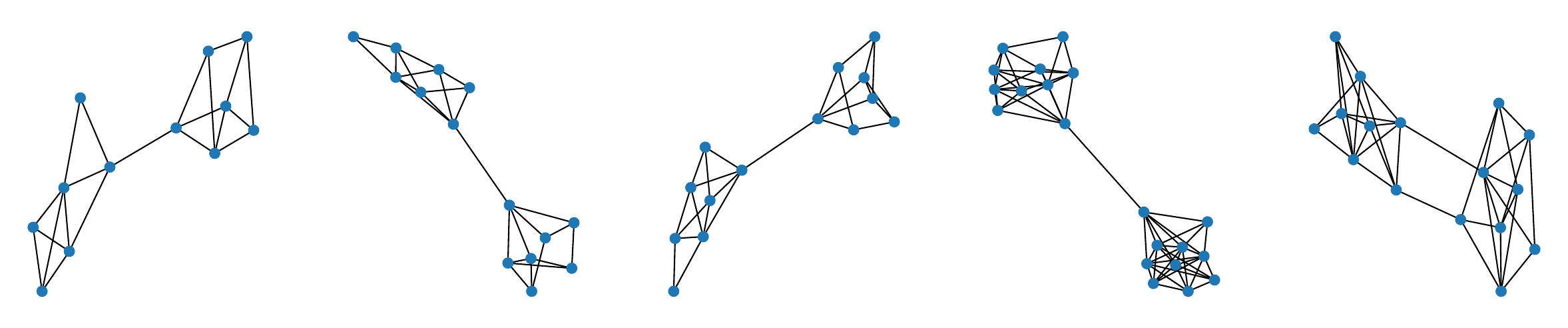}
        \caption{Data}
    \end{subfigure}

    \caption{Community samples.}\label{fig:comm_samples}
\end{figure}

%% file: main-paper/fig_grids/egosm.tex
\begin{figure}
    \centering
    \begin{subfigure}[t]{0.45\textwidth}
        \centering
        \includegraphics[width=\textwidth]{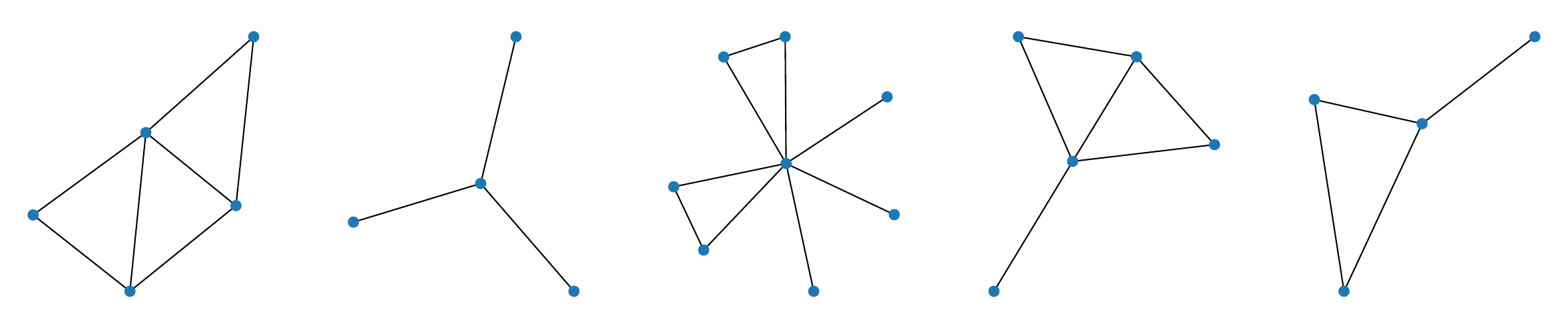}
        \caption{Seq-big FLAGG}
    \end{subfigure}
    \hspace{1cm}
    \begin{subfigure}[t]{0.45\textwidth}
        \centering
        \includegraphics[width=\textwidth]{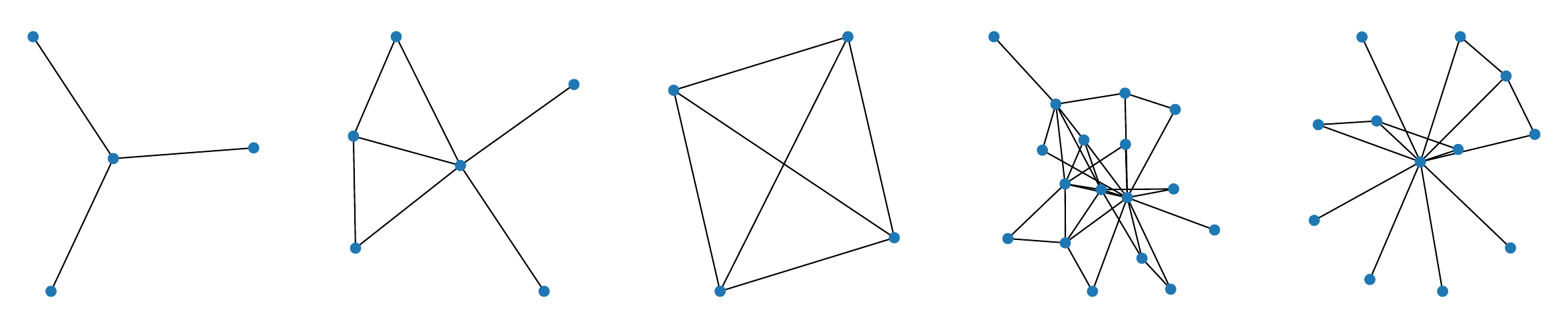}
        \caption{Properties FLAGG}
    \end{subfigure}

    \begin{subfigure}[t]{0.45\textwidth}
        \centering
        \includegraphics[width=\textwidth]{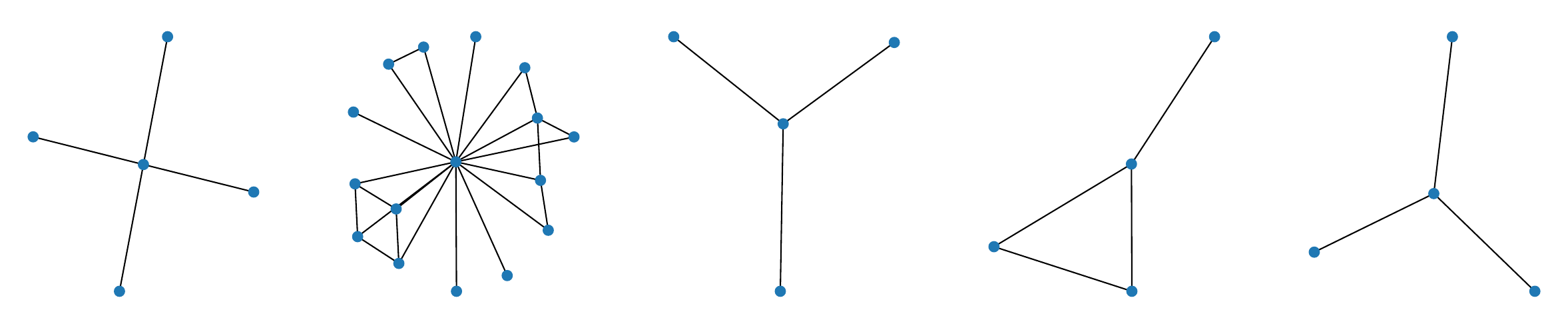}
        \caption{Seq-small FLAGG}
    \end{subfigure}
    \hspace{1cm}
    \begin{subfigure}[t]{0.45\textwidth}
        \centering
        \includegraphics[width=\textwidth]{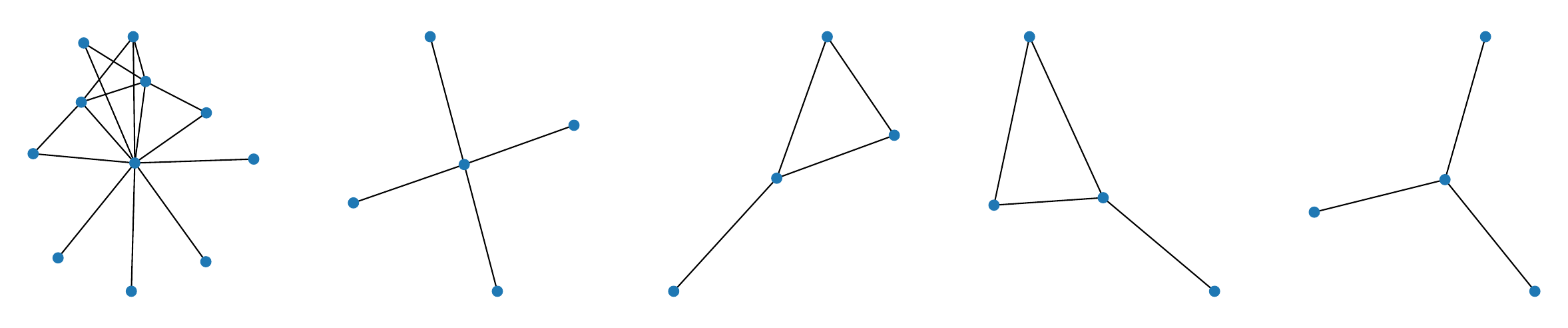}
        \caption{Data}
    \end{subfigure}

    \caption{Ego-Small samples.}\label{fig:egosm_samples}
\end{figure}

%% file: main-paper/fig_grids/enz.tex
\begin{figure}
    \centering
    \begin{subfigure}[t]{0.45\textwidth}
        \centering
        \includegraphics[width=\textwidth]{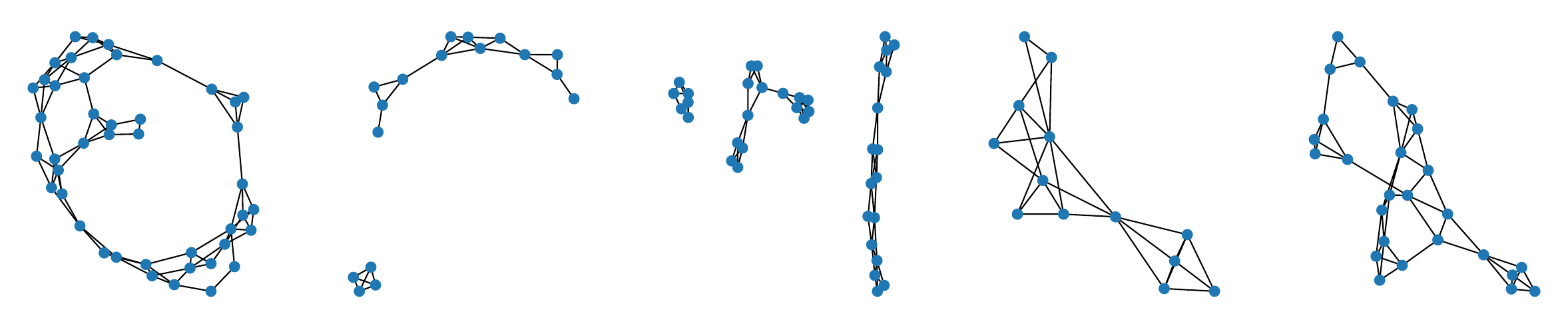}
        \caption{Seq-big FLAGG}
    \end{subfigure}
    \hspace{1cm}
    \begin{subfigure}[t]{0.45\textwidth}
        \centering
        \includegraphics[width=\textwidth]{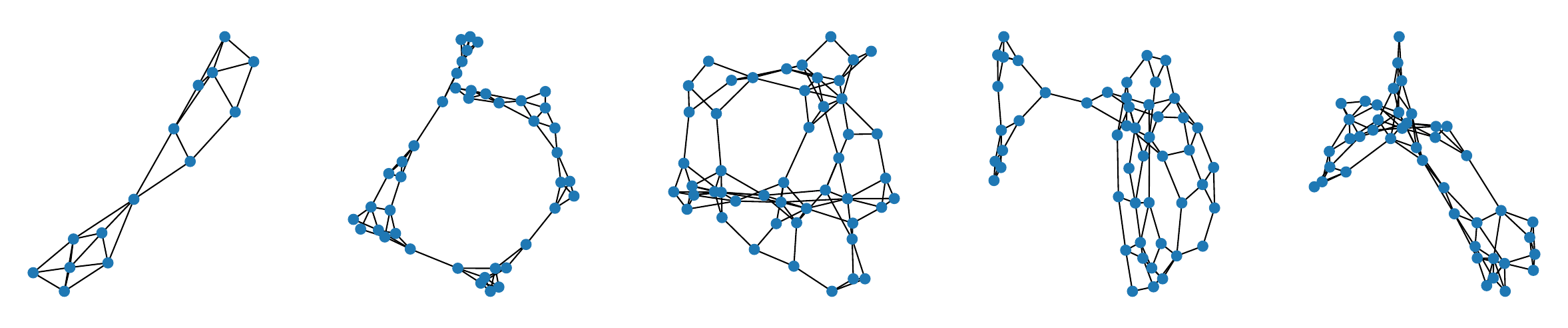}
        \caption{Properties FLAGG}
    \end{subfigure}

    \begin{subfigure}[t]{0.45\textwidth}
        \centering
        \includegraphics[width=\textwidth]{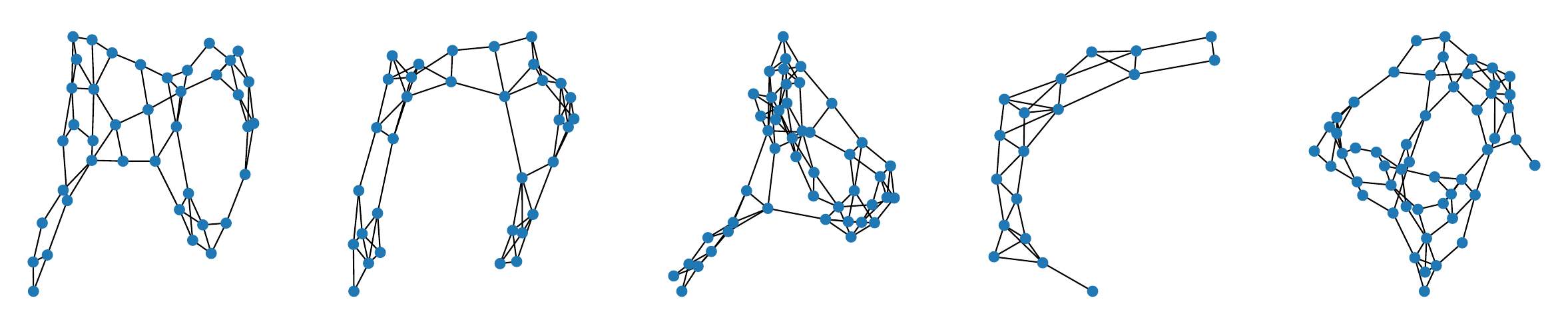}
        \caption{Seq-small FLAGG}
    \end{subfigure}
    \hspace{1cm}
    \begin{subfigure}[t]{0.45\textwidth}
        \centering
        \includegraphics[width=\textwidth]{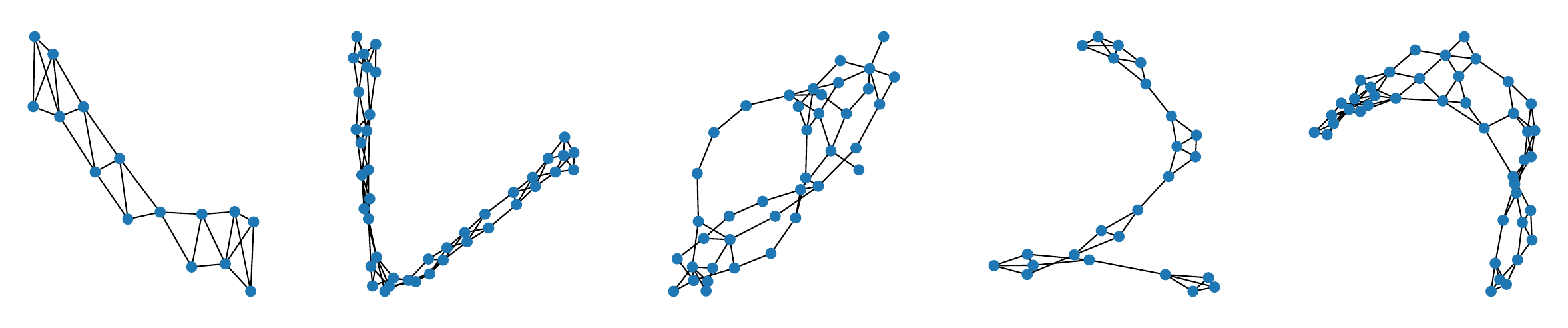}
        \caption{Data}
    \end{subfigure}

    \caption{Enzymes samples.}\label{fig:enz_samples}
\end{figure}

%% file: main-paper/fig_grids/ego.tex
\begin{figure}
    \centering
    \begin{subfigure}[t]{0.45\textwidth}
        \centering
        \includegraphics[width=\textwidth]{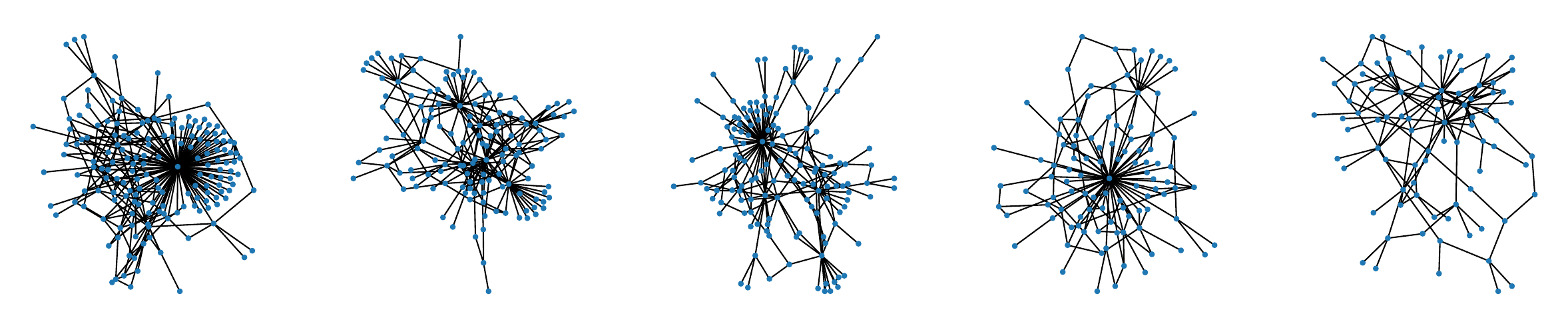}
        \caption{Seq-big FLAGG}
    \end{subfigure}
    \hspace{1cm}
    \begin{subfigure}[t]{0.45\textwidth}
        \centering
        \includegraphics[width=\textwidth]{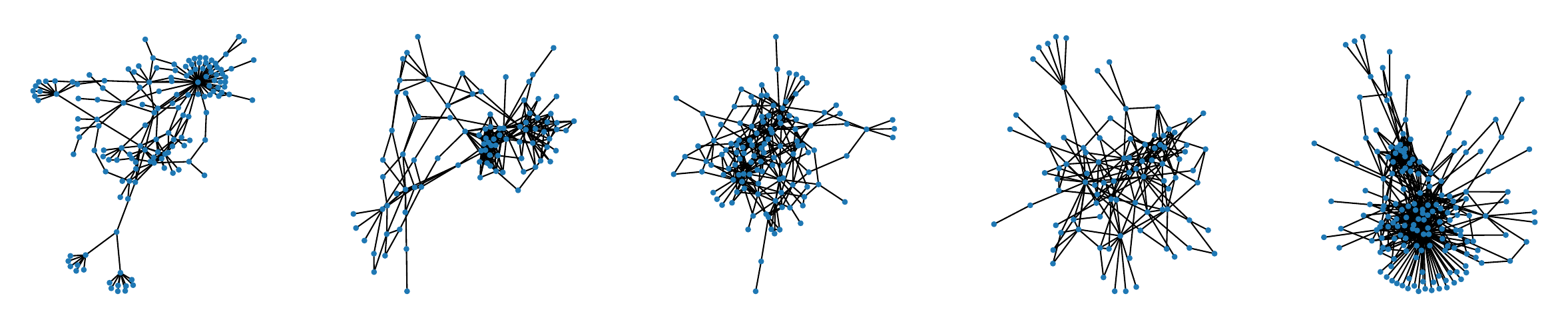}
        \caption{Properties FLAGG}
    \end{subfigure}

    \begin{subfigure}[t]{0.45\textwidth}
        \centering
        \includegraphics[width=\textwidth]{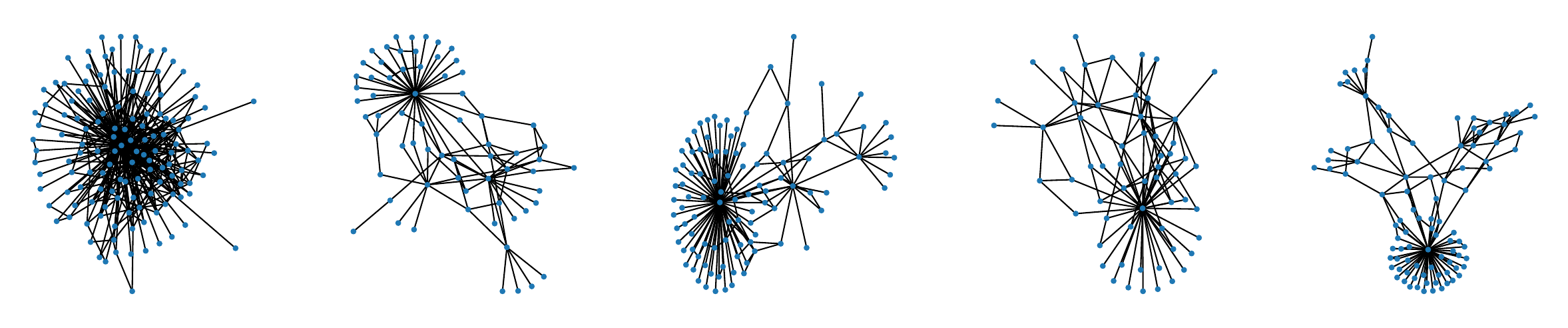}
        \caption{Seq-small FLAGG}
    \end{subfigure}
    \hspace{1cm}
    \begin{subfigure}[t]{0.45\textwidth}
        \centering
        \includegraphics[width=\textwidth]{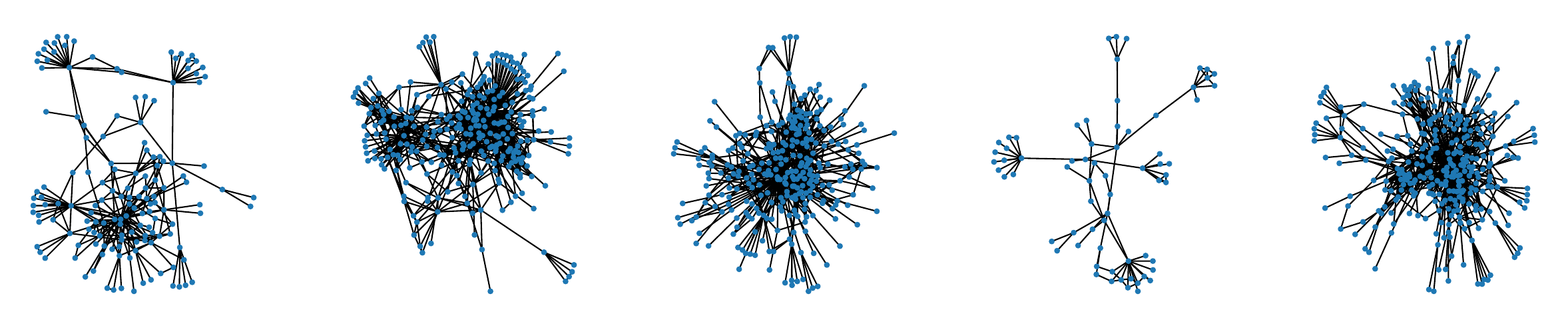}
        \caption{Data}
    \end{subfigure}

    \caption{Ego samples.}\label{fig:ego_samples}
\end{figure}

%% file: main-paper/fig_grids/qm9.tex
\begin{figure}
    \centering

    \begin{minipage}[c]{0.45\textwidth}
        \begin{subfigure}[t]{\textwidth}
            \centering
            \includegraphics[width=\textwidth]{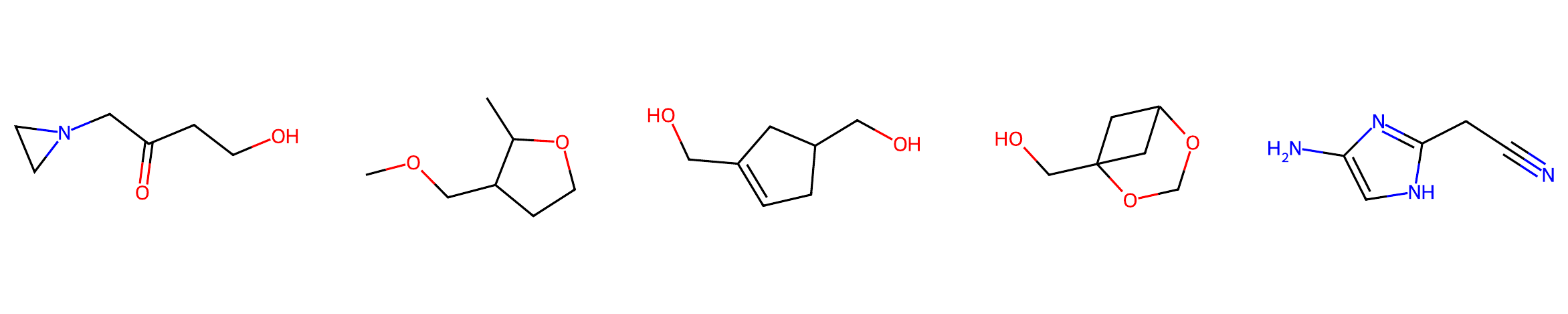}
            \caption{Seq-big FLAGG}
        \end{subfigure}

        \begin{subfigure}[t]{\textwidth}
            \centering
            \includegraphics[width=\textwidth]{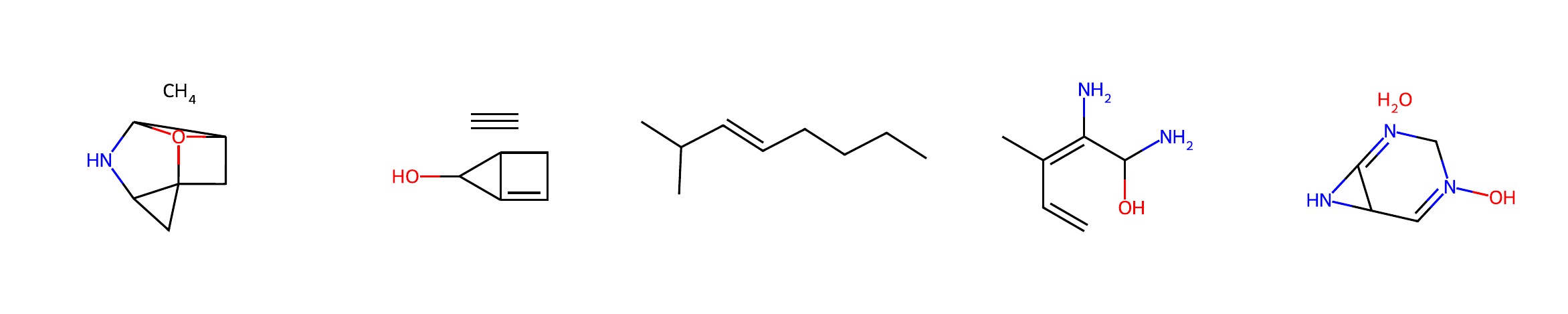}
            \caption{Seq-small FLAGG}
        \end{subfigure}
        
    \end{minipage}
    \hspace{1cm}
    \begin{subfigure}[c]{0.45\textwidth}
        \centering
        \includegraphics[width=\textwidth]{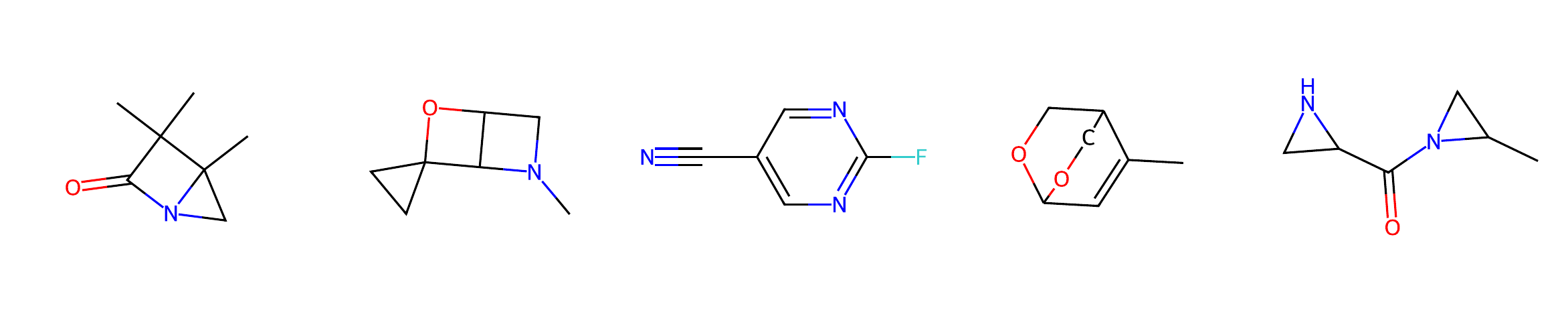}
        \caption{Data}
    \end{subfigure}

    \caption{QM9 samples.}\label{fig:qm9_samples}
\end{figure}

%% file: main-paper/fig_grids/zinc.tex
\begin{figure}
    \centering
    \begin{minipage}[c]{0.45\textwidth}
        \begin{subfigure}[t]{\textwidth}
            \centering
            \includegraphics[width=\textwidth]{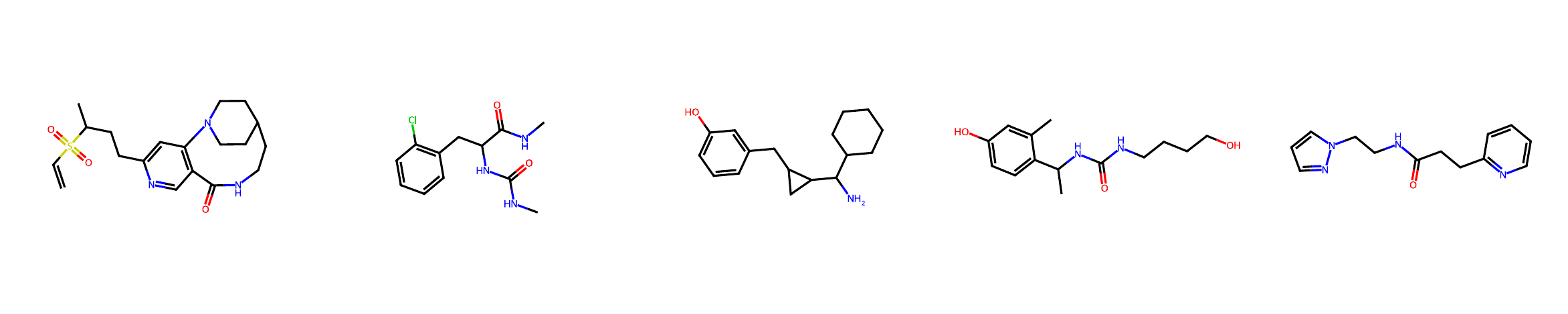}
            \caption{Seq-big FLAGG}
        \end{subfigure}

        \begin{subfigure}[t]{\textwidth}
            \centering
            \includegraphics[width=\textwidth]{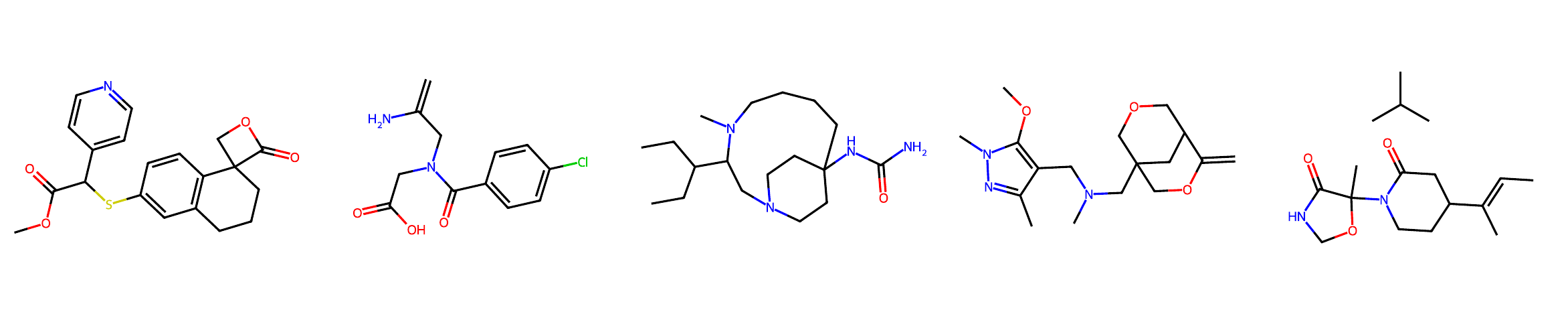}
            \caption{Seq-small FLAGG}
        \end{subfigure}
        
    \end{minipage}
    \hspace{1cm}
    \begin{subfigure}[c]{0.45\textwidth}
        \centering
        \includegraphics[width=\textwidth]{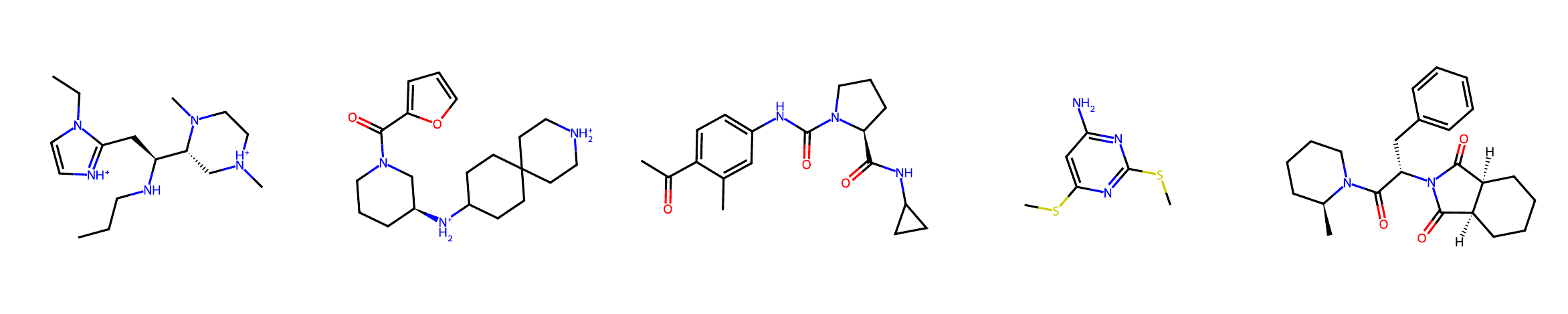}
        \caption{Data}
    \end{subfigure}

    \caption{Zinc samples.}\label{fig:zinc_samples}
\end{figure}

%% file: main-paper/tables/sampling_time.tex
\begin{table*}[t]
    \centering

    \begin{subtable}[t]{0.60\textwidth}
        \centering
        \begin{tabular}{*{3}{c}}
            \toprule
            Method & QM9 & ZINC250k\\
            \midrule
            seq-small & $122.616_{21.507}$ & $295.300_{18.903}$ \\
            seq-big & $96.017_{16.266}$ & $151.949_{6.764}$ \\
            \bottomrule
        \end{tabular}
        \caption{Sampling time (in minutes) on the molecular datasets.}
    \end{subtable}
    \hspace{0.2cm}
    \begin{subtable}[t]{0.35\textwidth}
        \centering
        \begin{tabular}{*{2}{c}}
            \toprule
            Method & Cora\\
            \midrule
            sparse & $45.865_{1.473}$ \\
            \bottomrule
        \end{tabular}
        \caption{Sampling time (in minutes) on the Cora dataset.}
    \end{subtable}
    
    \vspace{0.5cm}

    \begin{subtable}[t]{1.\textwidth}
        \centering
        \begin{tabular}{*{5}{c}}
            \toprule
            Method & Community & Ego-small & Enzymes & Ego \\
            \midrule
            seq-small & $70.710_{2.659}$ & $54.071_{6.470}$ & $38.831_{9.527}$ & $283.899_{19.988}$ \\
            seq-big & $22.013_{2.058}$ & $30.268_{2.003}$ & $18.840_{5.091}$ & $67.213_{7.579}$ \\
            properties & $0.170_{0.024}$ & $0.103_{0.005}$ & $0.118_{0.003}$ & $1.259_{0.130}$ \\
            \bottomrule
        \end{tabular}
        \caption{Sampling time (in minutes) on the generic graph datasets.}
    \end{subtable}

    \vspace{0.2cm}

    \caption{Sampling time of the test batch of molecular graphs (a), Cora (b) and generic graphs (c) averaged over 3 runs. Values are reported as $\text{mean}_{\text{std}}$.}\label{tab:sampling_time}
\end{table*}

%% file: main-paper/fig_grids/tradeoff.tex
\begin{figure}[t]
    \centering
    \includegraphics[width=0.95\textwidth]{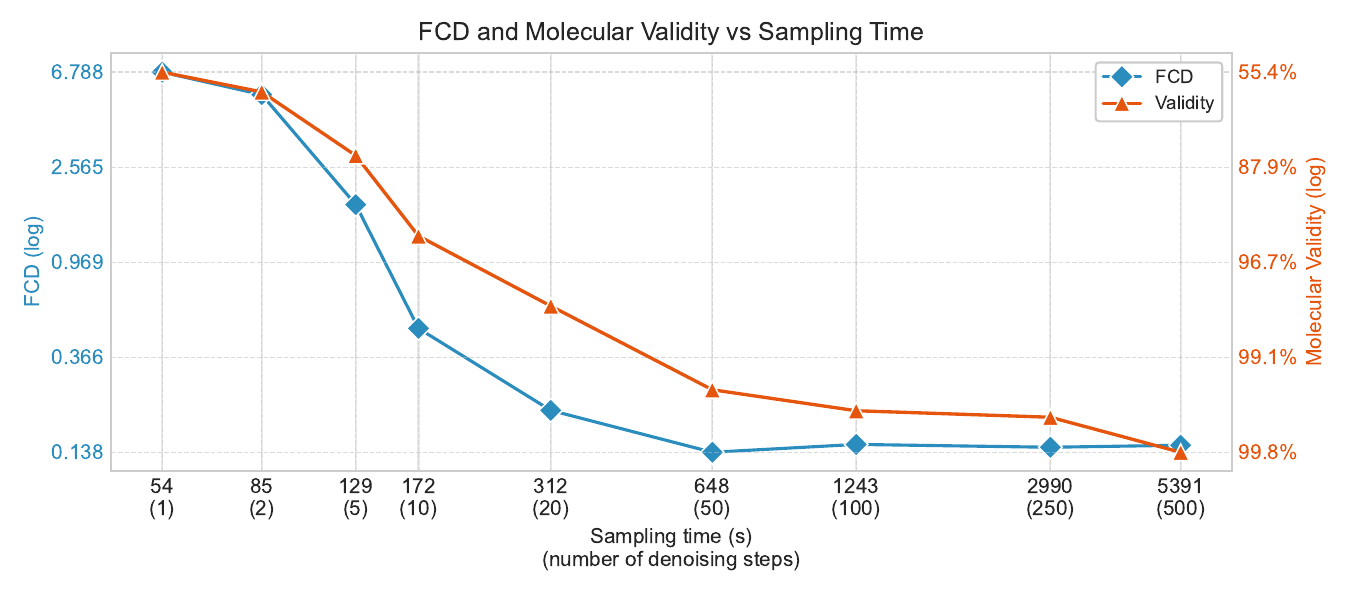}%
    \caption{Performance vs. sampling time of Vanilla FLAGG with different numbers of denoising steps on the QM9 data set, using block sizes $\{1,4\}$. Performance is evaluated in terms of FCD (blue line) and Molecular Validity (orange line). Vertical lines represent runs with different numbers of denoising steps, shown at the center of the line. Each point is the average across 3 different models trained with different seeds. All scales are logarithmic.}\label{fig:perf_vs_time}
\end{figure}

%% file: main-paper/tables/ablation.tex
\begin{table}[t]
    \centering

    \begin{tabular}{*{6}{c}}
        \toprule
        Method & Valid (\%)$\uparrow$ & NSPDK$\downarrow$ & FCD$\downarrow$ & Unique (\%)$\uparrow$ & Novel (\%)$\uparrow$\\
        \midrule
        BFS & $\underline{99.69}_{0.17}$ & $\mathbf{3.05\text{e-}4}_{0.36\text{e-}4}$ & $\mathbf{0.144}_{0.010}$ & $96.96_{0.37}$ & $76.05_{0.91}$ \\
        DFS & $99.65_{0.15}$ & $\underline{3.24\text{e-}4}_{0.34\text{e-}4}$ & $\underline{0.153}_{0.018}$ & $97.13_{0.13}$ & $77.52_{1.20}$ \\
        Random & $\mathbf{99.72}_{0.14}$ & $3.42\text{e-}4_{0.38\text{e-}4}$ & $0.253_{0.043}$ & $96.95_{0.21}$ & $77.80_{1.40}$ \\
        \bottomrule
    \end{tabular}

    \caption{Ablation study results for node orderings on the molecule generation task on QM9, averaged over 3 runs. Best results are in bold, and the second best are underlined. Values are reported as $\text{mean}_{\text{std}}$.}\label{tab:ablation}
\end{table}

%% file: main-paper/tables/halting_base.tex
\begin{table}[t]
    \centering

    \begin{tabular}{*{7}{c}}
        \toprule
        Blocks & Insertion & Deg.$\downarrow$ & Clus.$\downarrow$ & Spec.$\downarrow$ & GIN$\downarrow$\\
        \midrule
        \multirow{2}{*}{ \{1, 3\} } & Learned & $\mathbf{0.048}_{0.009}$ & $\mathbf{0.043}_{0.02}$ & $0.029_{0.003}$ & $\mathbf{0.019}_{0.001}$ \\
        & Empirical & $0.050_{0.007}$ & $0.044_{0.03}$ & $\mathbf{0.025}_{0.006}$ & $\mathbf{0.019}_{0.002}$ \\
        \midrule
        \multirow{2}{*}{ \{1, 2, 8\} } 
        & Learned & $\mathbf{0.062}_{0.012}$ & $\mathbf{0.070}_{0.034}$ & $0.041_{0.010}$ & $0.048_{0.025}$ \\
        & Empirical & $0.067_{0.017}$ & $0.072_{0.042}$ & $\mathbf{0.031}_{0.007}$ & $\mathbf{0.037}_{0.015}$ \\
        \bottomrule
    \end{tabular}

    \caption{Comparison of FLAGG with an ``Empirical'' baseline, sampling the number of nodes from the empirical distribution, averaged over 3 runs on the Enzymes dataset. Our method, sampling block sizes and the halting signal through learned neural networks, is identified as ``Learned''. Best results are in bold, and the second best are underlined. Values are reported as $\text{mean}_{\text{std}}$.}\label{tab:halting_baseline}
\end{table}

%% file: main-paper/fig_grids/halt_grid.tex
\begin{figure}[p]
    \centering
    \includegraphics[width=\textwidth]{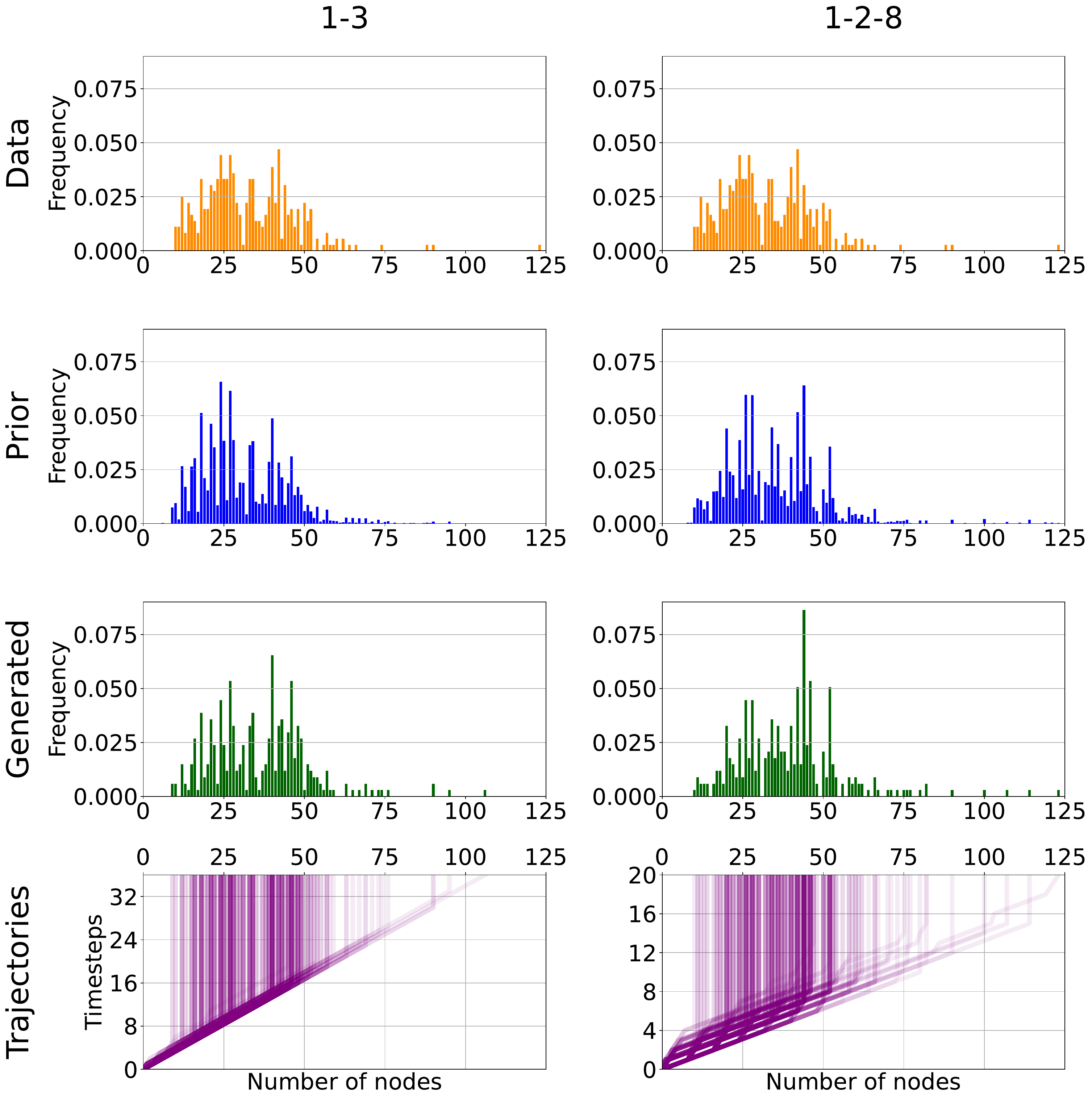}

    \caption{Analysis of the learned insertion policy on the Enzymes dataset. The two columns represent our two variants of FLAGG, trained with different sets of block sizes. The ``Data'' row shows the empirical distribution of the number of nodes in the training set of Enzymes. The ``Prior'' row shows the average distribution on nodes, given by the prior probability of halting at each timestep. The ``Generated'' row shows the number of nodes actually sampled. The ``Trajectories'' row shows the paths taken by each graph during its generative trajectory, where the y-axis represents time flowing from bottom to top. The intensity of the color represents the number of overlapping trajectories. Thus, the latter two rows are aligned, with the ``Generated'' histogram being an alternative representation of the x-axis of the ``Trajectories'' at the final timestep.}
    \label{fig:halt_grid}
\end{figure}

%% file: main-paper/fig_grids/blocks_grid.tex
\begin{figure}[p]
    \centering
    \begin{subfigure}[t]{0.45\textwidth}
        \centering
        \includegraphics[width=\textwidth]{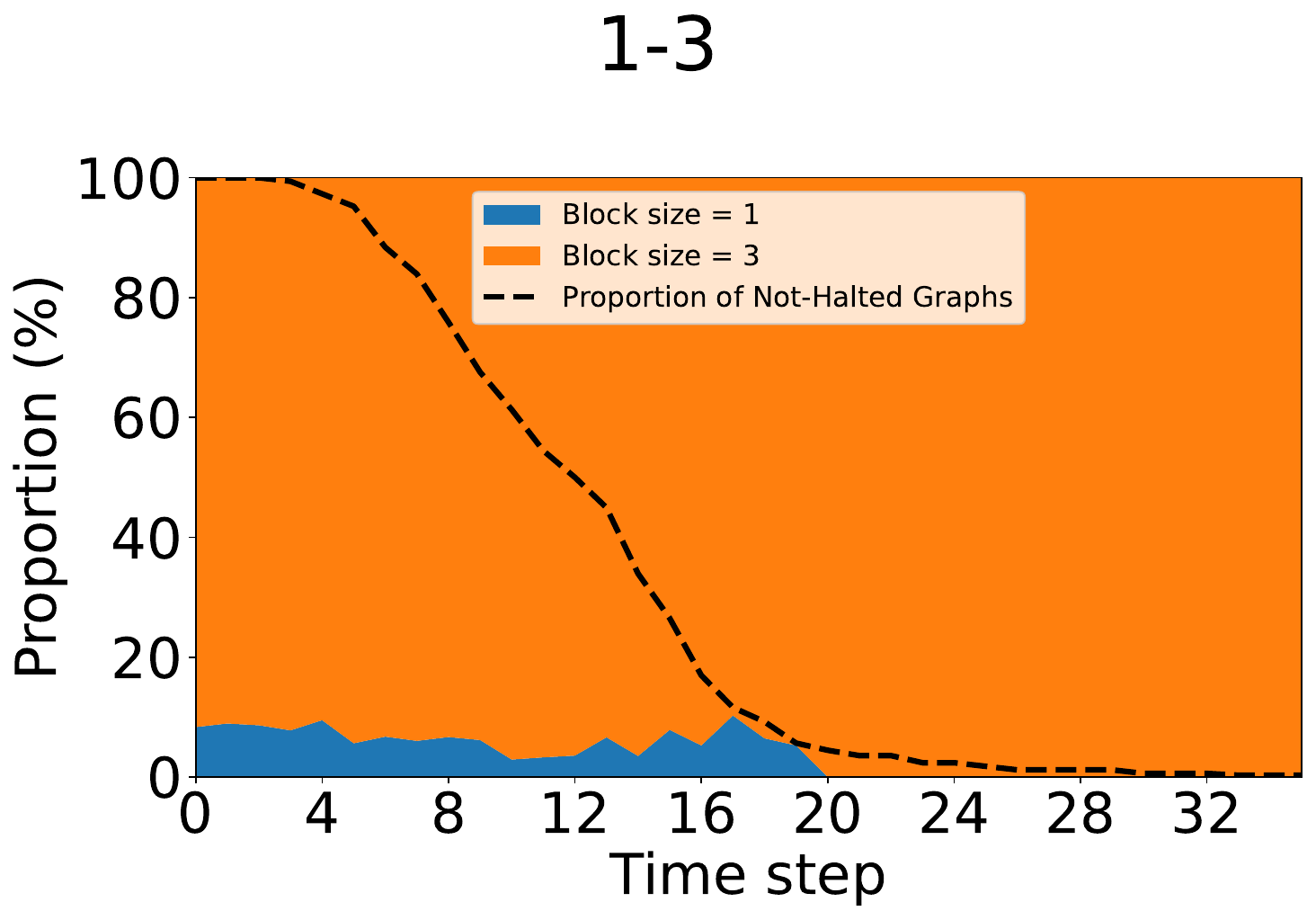}
    \end{subfigure}
    \hspace{0.5cm}
    \begin{subfigure}[t]{0.45\textwidth}
        \centering
        \includegraphics[width=\textwidth]{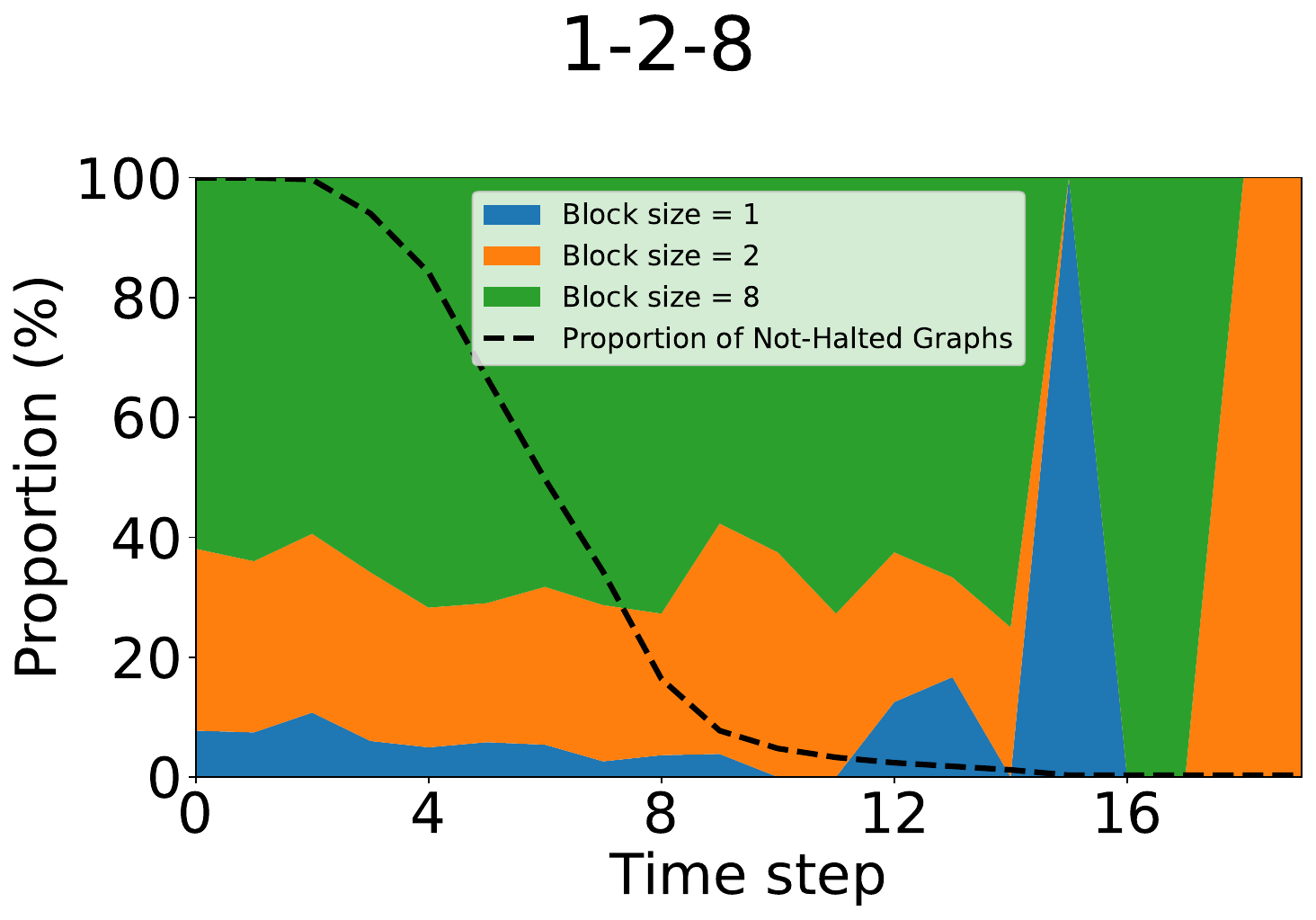}
    \end{subfigure}

    \vspace{1.0cm}

    \begin{subfigure}[t]{0.45\textwidth}
        \centering
        \includegraphics[width=\textwidth]{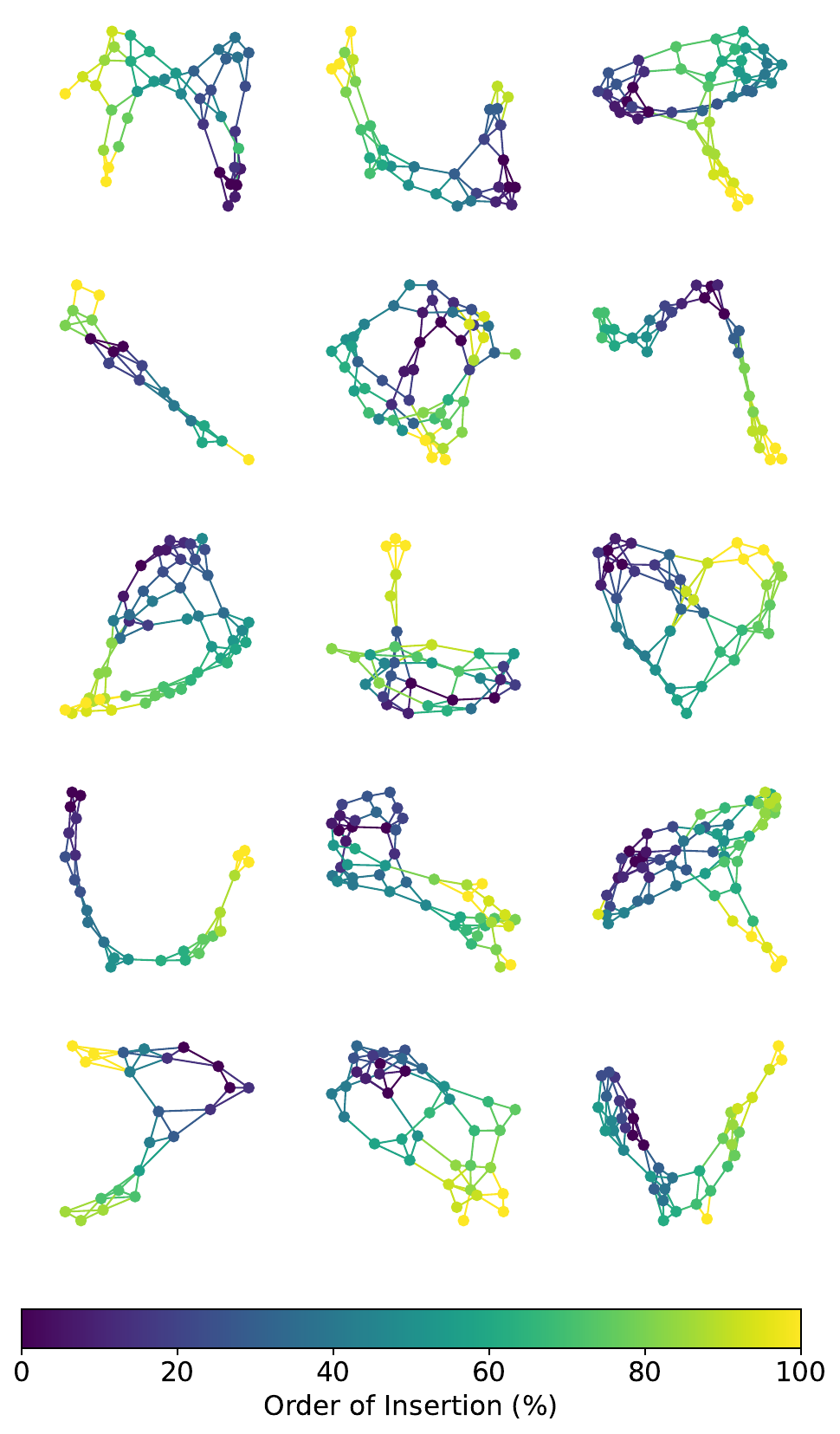}
    \end{subfigure}
    \hspace{0.5cm}
    \begin{subfigure}[t]{0.45\textwidth}
        \centering
        \includegraphics[width=\textwidth]{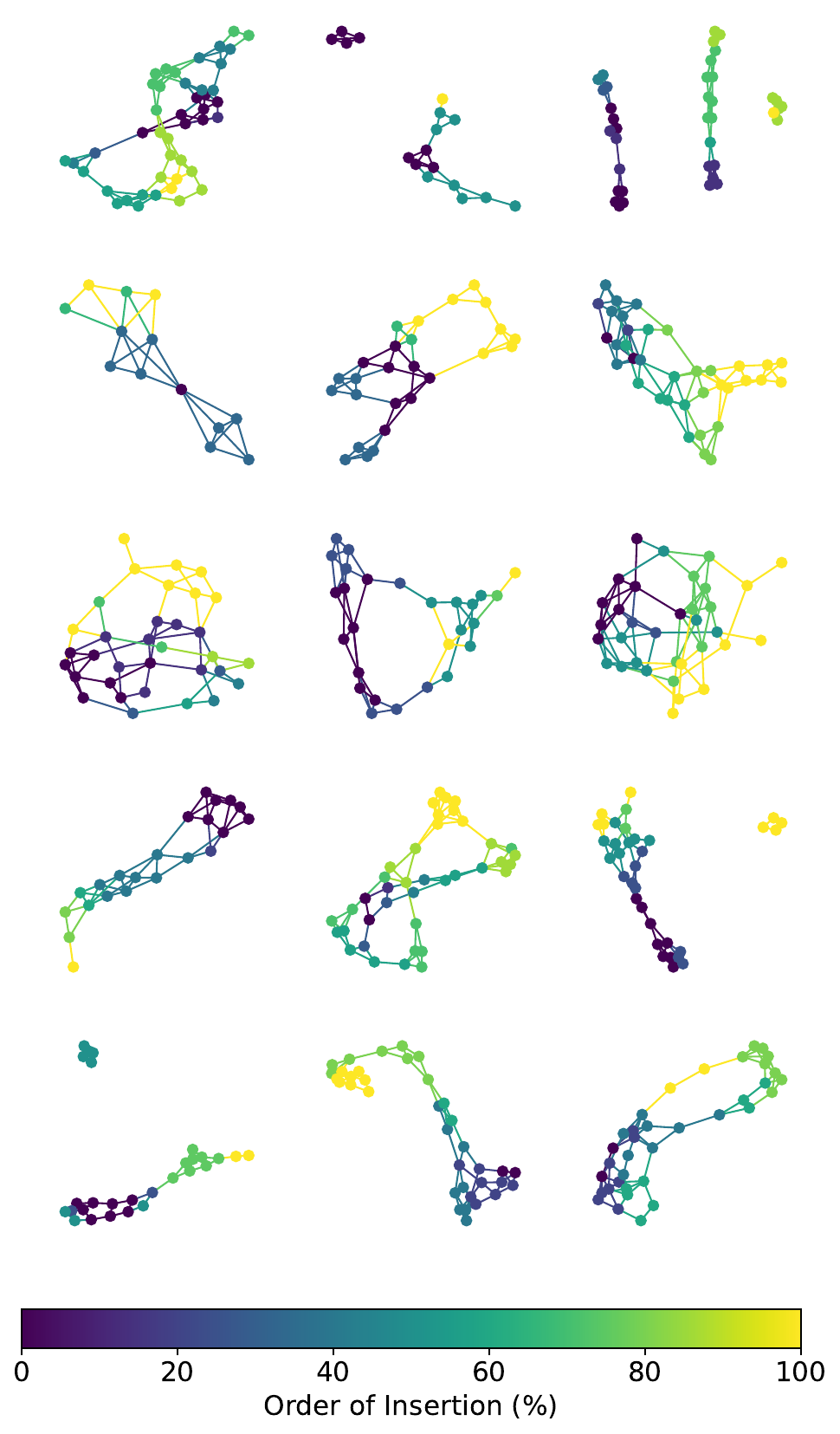}
    \end{subfigure}

    \caption{Top row: proportion (y-axis) of block sizes used during generation at each time step (x-axis) on the running batch of graphs, where the black dashed line indicates its size. Bottom row: example of generated graphs with node and edge colors indicating the \% time step at which they were generated.}
    \label{fig:blocks_grid}

\end{figure}

%% file: main-paper/conclusion.tex
\section{Conclusion}
In this paper we propose the Flexible Autoregressive Graph Generation (FLAGG) framework, allowing the harnessing of the generative power of one-shot models in a sequential setting. We improved over our preliminary model IFH, showing that generating with blocks is feasible and leads to high-quality graph generation. We also showed that thanks to the flexibility of FLAGG, we can define several variants, each with its advantages. For example, we can make the model consider subsets of nodes instead of the whole graph to generate large graphs. For boosting the learning of specific properties, we can make those explicit through the factorization trick of FLAGG. Finally, we observed that, in our setup, the model is not very sensible to the choice of the block sizes, but we require further research in this direction. We believe that FLAGG, with its freedom of design, is a promising framework for spurring new interest in autoregressive graph generation.

%% file: main-paper/acknowledgment.tex
\acks{
We acknowledge the financial support of (i) the PNRR project FAIR --- Future AI Research (PE00000013), under the NRRP MUR program funded by the NextGenerationEU; (ii) the Italian Ministry of University and Research (MUR) under the PRIN program – Progetti di Rilevante Interesse Nazionale – PRIN 2022 (Secretary-General's Decree No. 1401 of 18/09/2024) – CUP C53C24000770006, project title ``DEEP-GRAPH: Design and Theory of Deep Graph Learning''.
}

%% file: appendix/proofs.tex
\section{Proofs}\label{app:proofs}

\subsection{Proof of the Variational Lower Bound~\ref{eq:ifh_loss}}
\begin{proof}
Recall the notation in Section~\ref{sec:method/defs} of the main paper. To simplify the notation we consider $\mathcal{F}(G)$ as the set of any forward removal sequence of $G$. To prove loss~\eqref{eq:ifh_loss}, we start from the prior distribution of the model:
\begin{align}
p_{\theta ,\phi }& (\gG_{0} ) = \sum\limits _{\gG_{1:T} \in \mathcal{F} (\gG_{0})} p_{\theta ,\phi } (\gG_{0:T}) \nonumber\\
 & =\sum\limits _{\gG_{1:T} \in \mathcal{F} (\gG_{0})} p_{\theta ,\phi } (\gG_{0:T})\frac{q(\gG_{1:T} |\gG_{0} )}{q(\gG_{1:T} |\gG_{0} )} \label{proof:imp_sampl} \\
 & =\sum\limits _{\gG_{1:T} \in \mathcal{F} (\gG_{0})} q(\gG_{1:T} |\gG_{0} )p_{\theta} (\gG_{T} )\frac{p_{\theta ,\phi } (\gG_{0:T-1} |\gG_{T} )}{q(\gG_{1:T} |\gG_{0} )}\nonumber\\
 & =\sum\limits _{\gG_{1:T} \in \mathcal{F} (\gG_{0})} q(\gG_{1:T} |\gG_{0} )p_{\theta} (\gG_{T} )\prod _{t=1}^{T}\frac{p_{\theta ,\phi } (\gG_{t-1} |\gG_{t} )}{q(\gG_{t} |\gG_{t-1})} \label{proof:markov}\\
 & =\sum\limits _{\gG_{1:T} \in \mathcal{F} (\gG_{0})} q(\gG_{1:T} |\gG_{0} )\frac{p_{\theta} (\gG_{T} )}{q( \gG_{T} |\gG_{0})} p_{\theta ,\phi } (\gG_{0} |\gG_{1} )\prod _{t=2}^{T}\frac{p_{\theta ,\phi } (\gG_{t-1} |\gG_{t} )}{q(\gG_{t-1} |\gG_{t} ,\gG_{0})} \nonumber\\
 & =\sum\limits _{\gG_{1:T} \in \mathcal{F} (\gG_{0})} q(\gG_{1:T} |\gG_{0} )\frac{p_{\theta} (\gG_{T} )}{q(\gG_{T} |\gG_{0})} p_\phi (n_{0} |\gG_{1} ) p_{\theta} (\gG_{0} |n_{0} ,\gG_{1} ) \nonumber\\
 &\quad\quad\quad\cdot \prod _{t=2}^{T}\frac{p_\phi (n_{t-1} |\gG_{t} )}{q( n_{t-1} |\gG_{t} ,\gG_{0})}\frac{p_{\theta} (\gG_{t-1} |n_{t-1} ,\gG_{t} )}{q(\gG_{t-1} |n_{t-1} ,\gG_{t} ,\gG_{0})}\nonumber\\
 & =\sum\limits _{\gG_{1:T} \in \mathcal{F} (\gG_{0})} q(\gG_{1:T} |\gG_{0} )\frac{p_{\theta} (\gG_{T} )}{q(\gG_{T} |\gG_{0})} p_\phi (r_1 |\gG_{1} ) p_{\theta} (\gW_1 |r_1 ,\gG_{1} )\prod _{t=2}^{T}\frac{p_\phi (r_t |\gG_{t} )}{q( r_t |\gG_{t} ,\gG_{0})}\frac{p_{\theta} (\gW_t |r_t ,\gG_{t} )}{q( \gW_t |r_t ,\gG_{t} ,\gG_{0})}\nonumber\\
 & =\sum\limits _{\gG_{1:T} \in \mathcal{F} (\gG_{0})} q(\gG_{1:T} |\gG_{0} ) p_\phi (r_1 |\gG_{1} ) p_{\theta} (\gW_1 |r_1 ,\gG_{1} )\prod _{t=2}^{T}\frac{p_\phi (r_t |\gG_{t} )}{q( r_t |\gG_{t} ,\gG_{0})}\frac{p_{\theta} (\gW_t |r_t ,\gG_{t} )}{q( \gW_t |r_t ,\gG_{t} ,\gG_{0})}.\nonumber
\end{align}
In step~\ref{proof:imp_sampl} we used importance sampling to change the variable over which the expectation is computed, and step~\ref{proof:markov} where we factorized the probabilities over sequences with the definition of removal and insertion processes (respectively Equations~\eqref{eq:remv_proc}~and~\eqref{eq:ins_proc}). 
The Variational Upper Bound (VUB) is found computing the negative log-likelihood and applying the Jensen Inequality:
\begin{align*}
    \E_{q(\gG_{0})}&[-\log p_{\theta ,\phi } (\gG_{0} )] \leq 
    \E_{q(\gG_{0})}\Bigg[\sum\limits_{t=2}^{T}\KL \big(q(r_t|\gG_t,\gG_0) \Vert p_{\phi}(r_t|\gG_t)\big)-\E_{q(\gG_1|\gG_0)}\left[\log p_\phi(r_{1}|\gG_1)\right]+ \nonumber\\
	&+\sum\limits_{t=2}^{T}\KL \big(q(\gW_t|r_t,\gG_t,\gG_0) \Vert p_{\theta}(\gW_t|r_t,\gG_t)\big)-\E_{q(\gG_1|\gG_0)}\left[\log p_\theta(\gW_1|r_1,\gG_1)\right]
	\Bigg]
\end{align*}
\end{proof}
\subsection{Proof of Equation~\ref{eq:bin_tsteps_num}}

\begin{proof}
Let's prove this by induction. Consider the simple case for $n_{1}$:
\begin{equation*}
q(n_{1}|n_{0})=B(n_{1};n_0,\pi_{1})
\end{equation*}
with $\pi_{1}=1-q_{1}$. This is true by the definition of binomial transitions in Equation~\eqref{eq:bin_next_num}.

Now, assume the property is true for $t-1$, that is, $n_{t-1}|n_{0}$ is a Binomial random variable $B(n_{t-1};n_0,\pi_{t-1})$. We know that $n_{t}|n_{t-1}$ is also a Binomial, and has the same distribution as $n_{t}|n_{t-1},n_{0}$ due to the Markov property. Let's recall what their distribution and parameters are:
\begin{gather*}
    n_{t}|n_{t-1},n_{0}\sim B(n_{t}; n_{t-1}|n_{0},1-q_{t}) \\
n_{t-1}|n_{0}\sim B(n_{t-1}; n_{0},\pi_{t-1}) \\
\text{with} \pi_{t-1}=\prod\limits_{k=1}^{t-1}(1-q_{k})
\end{gather*}
It can be proven that a Binomial r.v. conditioned on a Binomial r.v. is still a Binomial r.v. with success probability the product of the two success probabilities, and number of experiments the same as the conditioning binomial. From this fact $n_{t}|n_{0}$ is a Binomial r.v.:
\begin{gather*}
    n_{t}|n_{0}\sim B(n_t; n_{0},\pi_{t}) \\
    \pi_{t}=(1-q_{t})\pi_{t-1}=\prod\limits_{k=1}^{t}(1-q_{k})
\end{gather*}
\end{proof}

\subsection{Proof of Equation~\ref{eq:bin_post_num}}

\begin{proof}
Let's compute the posterior probability of the binomial removals:
\begin{align*}
q(n_{t-1}|&n_{t} ,n_{0} ) = \\
 & = q(n_{t} |n_{t-1} )\frac{q(n_{t-1} |n_{0} )}{q(n_{t} |n_{0} )}\\
 & =\frac{n_{t-1} !}{n_{t} !(n_{t-1} -n_{t} )!} (1-q_{t})^{n_{t}} q_{t}^{n_{t-1} -n_{t}} \frac{\frac{n_{0} !}{n_{t-1} !(n_{0} -n_{t-1} )!} \pi _{t-1}^{n_{t-1}} (1-\pi _{t-1} )^{n_{0} -n_{t-1}}}{\frac{n_{0} !}{n_{t} !(n_{0} -n_{t} )!} \pi _{t}^{n_{t}} (1-\pi _{t} )^{n_{0} -n_{t}}}\\
 & =\frac{(n_{0} -n_{t} )!}{(n_{t-1} -n_{t} )!(n_{0} -n_{t-1} )!} \pi _{t-1}^{n_{t-1} -n_{t}} q_{t}^{n_{t-1} -n_{t}}\frac{(1-\pi _{t-1} )^{n_{0} -n_{t-1}}}{(1-\pi _{t} )^{n_{0} -n_{t}}}\\
 & =\frac{(n_{0} -n_{t} )!}{(n_{t-1} -n_{t} )!(n_{0} -n_{t} -( n_{t-1} -n_{t}) )!}\pi _{t-1}^{n_{t-1} -n_{t}} (1-\pi _{t-1} )^{n_{0} -n_{t-1}}\frac{q_{t}^{n_{t-1} -n_{t}}}{(1-\pi _{t} )^{n_{0} -n_{t}}}\\
 & =\binom{n_{0} -n_{t}}{n_{t-1} -n_{t}}\left( q_{t}\frac{\pi _{t-1}}{1-\pi _{t}}\right)^{n_{t-1} -n_{t}}\left( q_{t}\frac{1-\pi _{t-1}}{1-\pi _{t}}\right)^{n_{0} -n_{t-1}}\\
 &=\binom{n_{0} -n_{t}}{n_{0} -n_{t-1}}\left(\frac{1-\pi _{t-1}}{1-\pi _{t}}\right)^{n_{0} -n_{t-1}}\left( 1-\frac{1-\pi _{t-1}}{1-\pi _{t}}\right)^{n_{t-1} -n_{t}}
\end{align*}

Finally, by substituting the number of failures at step $t$: $r_{t}=n_{t-1}+n_{t}$ we get:
\begin{equation*}
    q(r_{t} |n_{t} ,n_{0} ) =\binom{n_{0} -n_{t}}{r_{t}}\left( q_{t}\frac{\pi _{t-1}}{1-\pi _{t}}\right)^{r_{t}}\left(\frac{1-\pi _{t-1}}{1-\pi _{t}}\right)^{n_{0} -n_{t} -r_{t}}
\end{equation*}
\end{proof}

\subsection{Proof of Equation~\ref{eq:cat_tsteps_num}}

\begin{proof}
Let's prove this by induction. Consider the simple case for $n_{1}$:
\begin{equation*}
q(n_{1}|n_{0})=q(r_1|n_{0})
=\frac{\prod_{d\in D}\binom{h( n_{0})[d]}{h(r_1)[d]}}{\binom{T}{1}}
=\frac{h( n_{0})[r_1]}{T}
\end{equation*}
where the product over denominations only one non-unit factor with $d=r_1$, because $h(r_1)[r_1]=1$ and $h(r_1)[d]=0$ for all other denominations, as $r_1$ is one of the possible choices in $D$. Now, assume the property is true for $t-1$, that is, $n_{t-1}|n_{0}$ is a Multivariate hypergeometric, that is:
\begin{equation*}
q(n_{t-1}|n_0)=\frac{\prod_{d\in D}\binom{h( n_{0})[d]}{h(\Delta n_{t-1})[d]}}{\binom{T}{t-1}}
\end{equation*}

Now, using the law of total probability:
\begin{align*}
&q(n_t|n_0)
= \sum_{n_{t-1}=n_t}^{n_0}q(n_t|n_{t-1})q(n_{t-1}|n_0) \\
&= \sum_{d\in D} q(n_t|n_t+d)q(n_t+d|n_0) \\
&= \sum_{d\in D} \frac{h(n_t+d)[d]}{T-t+1}
\frac{\prod_{d'\in D}\binom{h( n_{0})[d']}{h(n_0 - n_t - d)[d']}}{\binom{T}{t-1}} \\
&=\frac{1}{(T-t+1)\frac{T!}{(T-t+1)!(t-1)!}}\sum _{d\in D} h(n_{t} +d)[d] \prod _{d'\in D}\binom{h(n_{0} )[d']}{h(n_{0} -n_{t} -d)[d']} \\
&=\frac{1}{t}\frac{1}{\frac{T!}{(T-t)!t!}}\sum _{d\in D}( h(n_{t} )[d]+1) \frac{h(n_{0} )[d]!}{( h(n_{0} )[d]-h(n_{0} -n_{t} -d)[d]) !h(n_{0} -n_{t} -d)[d]!} \\
&\quad\quad\quad\cdot \prod _{d'\in D\setminus \{d\}}\binom{h(n_{0} )[d']}{h(n_{0} -n_{t} -d)[d']}\\
&=\frac{1}{t}\frac{1}{\binom{T}{t}}\sum _{d\in D}( h(n_{t} )[d]+1) \frac{h(n_{0} )[d]!}{( h(n_{0} )[d]-h(n_{0} -n_{t} )[d]+1) !( h(n_{0} -n_{t} )[d]-1) !}\\
&\quad\quad\quad\cdot \prod _{d'\in D\setminus \{d\}}\binom{h(n_{0} )[d']}{h(n_{0} -n_{t} )[d']}\\
&=\frac{1}{t}\frac{1}{\binom{T}{t}}\sum _{d\in D}( h(n_{t} )[d]+1) \frac{h(n_{0} )[d]!}{( h(n_{t} )[d]+1) !( h(n_{0} )[d]-h(n_{t} )[d]-1) !} \\
&\quad\quad\quad\cdot \prod _{d'\in D\setminus \{d\}}\binom{h(n_{0} )[d']}{h(n_{0} -n_{t} )[d']}\\
&=\frac{1}{t}\frac{1}{\binom{T}{t}}\sum _{d\in D} h(n_{0} -n_{t} )[d]\frac{h(n_{0} )[d]!}{h(n_{t} )[d]!( h(n_{0} )[d]-h(n_{t} )[d]) !}\prod _{d'\in D\setminus \{d\}}\binom{h(n_{0} )[d']}{h(n_{0} -n_{t} )[d']}\\
&=\frac{1}{t}\frac{1}{\binom{T}{t}}\sum _{d\in D} h(n_{0} -n_{t} )[d]\binom{h(n_{0} )[d]}{h(n_{0} -n_{t} )[d]} \prod _{d'\in D\setminus \{d\}}\binom{h(n_{0} )[d']}{h(n_{0} -n_{t} )[d']}\\
&=\frac{1}{t}\frac{1}{\binom{T}{t}}\sum _{d\in D} h(n_{0} -n_{t} )[d]\prod _{d'\in D}\binom{h(n_{0} )[d']}{h(n_{0} -n_{t} )[d']}\\
&=\frac{1}{t}\frac{\prod _{d'\in D}\binom{h(n_{0} )[d']}{h(n_{0} -n_{t} )[d']}}{\binom{T}{t}}\sum _{d\in D} h(n_{0} -n_{t} )[d]\\
&=\frac{1}{t}\frac{\prod _{d'\in D}\binom{h(n_{0} )[d']}{h(n_{0} -n_{t} )[d']}}{\binom{T}{t}} t\\
&=\frac{\prod _{d'\in D}\binom{h(n_{0} )[d']}{h(\Delta n_{t} )[d']}}{\binom{T}{t}}
\end{align*}

To reach the final statement we used the following facts:
\begin{itemize}
    \item $h(n+d)[d']=\begin{cases}
        h(n)[d]+1 & \text{for }d'=d\\
        h(n)[d'] & \text{otherwise}
    \end{cases}$
    \item $h(n_0)-h(n_t)=h(n_0-n_t)$
    \item by definition $\sum_{d\in D}h(n_0-n_t)[d]=t$
\end{itemize}
\end{proof}

\subsection{Proof of Equation~\ref{eq:cat_post_num}}

\begin{proof}
Let's compute the posterior:
\begin{align*}
q( n_{t-1} &|n_{0} ,n_{t}) = \frac{q( n_{t} |n_{t-1}) qp( n_{t-1} |n_{0})}{q( n_{t} |n_{0})} \\
&=\frac{h( n_{t-1})[ r_{t}]}{T-t+1}\frac{\prod _{d\in D}\frac{h( n_{0})[ d] !}{h( \Delta n_{t-1})[ d] ! h( n_{t-1})[ d]!}}{\frac{T!}{( t-1) !( T-t+1) !}}\left(\frac{\prod _{d\in D}\frac{h( n_{0})[ d] !}{h( \Delta n_{t})[ d] ! h( n_{t})[ d] !}}{\frac{T!}{t!( T-t) !}}\right)^{-1}\\
&=\frac{h( n_{t-1})[ r_{t}]( t-1) !( T-t+1) !}{t!( T-t) !( T-t+1)}\prod _{d\in D}\frac{h( \Delta n_{t})[ d] ! h( n_{t})[ d] !}{h( \Delta n_{t-1})[ d] ! h( n_{t-1})[ d] !}
\end{align*}
\begin{align*}
&=\frac{h( n_{t-1})[ r_{t}]}{t}\prod _{d\in D}\frac{h( \Delta n_{t})[ d] ! h( n_{t})[ d] !}{h( \Delta n_{t-1})[ d] ! h( n_{t-1})[ d] !}\\
&=\frac{h( n_{t-1})[ r_{t}]}{t}\prod _{d\in D}\frac{h( \Delta n_{t-1} +r_{t})[ d] ! h( n_{t})[ d] !}{h( \Delta n_{t-1})[ d] ! h( n_{t} +r_{t})[ d] !}\\
&=\frac{h( n_{t-1})[ r_{t}]}{t}\frac{h( \Delta n_{t-1} +r_{t})[ r_{t}] ! h( n_{t})[ r_{t}] !}{h( \Delta n_{t-1})[ r_{t}] ! h( n_{t} +r_{t})[ r_{t}] !}\prod _{d\in D\setminus \{r_{t}\}}\frac{h( \Delta n_{t-1})[ d] ! h( n_{t})[ d] !}{h( \Delta n_{t-1})[ d] ! h( n_{t})[ d] !}\\
&=\frac{h( n_{t-1})[ r_{t}]}{t}\frac{( h( \Delta n_{t-1})[ r_{t}] +1) ! h( n_{t})[ r_{t}] !}{h( \Delta n_{t-1})[ r_{t}] ! ( h( n_{t})[ r_{t}] +1) !}\\
&=\frac{h( n_{t})[ r_{t}] +1}{t}\frac{h( \Delta n_{t-1})[ r_{t}] +1}{h( n_{t})[ r_{t}] +1}\\
&=\frac{h( \Delta n_{t})[ r_{t}]}{t}
\end{align*}

Because $q(n_{t-1}|n_0,n_t)=q(r_t|n_0,n_t)$:
\begin{equation*}
q(r_t|n_0,n_t)=\frac{h( \Delta n_{t})[ r_{t}]}{t}
\end{equation*}

\end{proof}